\pdfoutput=1

\documentclass[11pt]{article}

\usepackage{acl}

\usepackage{times}
\usepackage{latexsym}

\usepackage[T1]{fontenc}

\usepackage[utf8]{inputenc}

\usepackage{microtype}

\usepackage{algorithm}
\usepackage{algorithmic}
\usepackage{array}
\usepackage{mathtools}
\usepackage{amsmath}
\usepackage{dsfont}
\usepackage{subfigure}
\usepackage{inconsolata}
\usepackage{graphicx}
\usepackage{multicol}
\usepackage{multirow}
\usepackage{xcolor}
\usepackage{colortbl}
\definecolor{Gray}{gray}{0.93}
\usepackage{amssymb}
\usepackage{amsthm}
\usepackage{enumitem}
\usepackage{bm}
\usepackage{booktabs}
\usepackage{units}
\usepackage{bbm}
\usepackage{pifont}
\usepackage{easyReview}
\newcommand{\modelname}{ImbaNID}
\newcommand{\taskname}{i-NID}
\newcommand{\dataset}{ImbaNID-Bench}
\newcommand\identity{1\kern-0.25em\text{l}}

\title{Towards Real-world Scenario:  Imbalanced New Intent Discovery}

\author{
Shun Zhang$^{\spadesuit\heartsuit}$, \quad
Chaoran Yan$^{\spadesuit}$
, \quad
Jian Yang$^{\spadesuit}$\thanks{Corresponding author.}, \quad
Jiaheng Liu$^{\spadesuit}$, \quad 
\textbf{Ying Mo}$^{\spadesuit}$, \quad \\
\textbf{Jiaqi Bai}$^{\spadesuit\heartsuit}$, \quad
\textbf{Tongliang Li}$^{\clubsuit}$, \quad
\textbf{Zhoujun Li}$^{\spadesuit\heartsuit}$ \quad
\\
${\spadesuit}$ State Key Laboratory of Complex \& Critical Software Environment, Beihang University \\ 
${\heartsuit}$ School of Cyber Science and Technology, Beihang University \\
${\clubsuit}$ Computer School, Beijing Information Science and Technology University \\
\texttt{\{shunzhang, ycr2345, jiaya, bjq,lizj\}@buaa.edu.cn} \quad \texttt{\{tonylingli\}@bistu.edu.cn} \quad
}

\begin{document}
\maketitle
\begin{abstract}
New Intent Discovery (NID) aims at detecting known and previously undefined categories of user intent by utilizing limited labeled and massive unlabeled data. 
Most prior works often operate under the unrealistic assumption that the distribution of both familiar and new intent classes is uniform, overlooking the skewed and long-tailed distributions frequently encountered in real-world scenarios. To bridge the gap, our work introduces the imbalanced new intent discovery (\taskname{}) task, which seeks to identify familiar and novel intent categories within long-tailed distributions. A new benchmark (\dataset{}) comprised of three datasets is created to simulate the real-world long-tail distributions. \dataset{} ranges from broad cross-domain to specific single-domain intent categories, providing a thorough representation of practical use cases. Besides, a robust baseline model \modelname{} is proposed to achieve cluster-friendly intent representations. It includes three stages: model pre-training, generation of reliable pseudo-labels, and robust representation learning that strengthens the model performance to handle the intricacies of real-world data distributions.
Our extensive experiments on previous benchmarks and the newly established benchmark demonstrate the superior performance of \modelname{} in addressing the \taskname{} task, highlighting its potential as a powerful baseline for uncovering and categorizing user intents in imbalanced and long-tailed distributions\footnote{\url{https://github.com/Zkdc/i-NID}}.
\end{abstract}

\section{Introduction}
New intent discovery (NID) has captured increasing attention due to its adaptability to the evolving user needs in open-world scenarios~\cite{gid,zero_shot_gid,alm,nid_industry_setting,idas}. NID methods generally follow a two-stage training process, including a knowledge transfer and a discovery stage. The prior knowledge is injected into the model via pre-training and then the discriminative representation is learned for known and novel intent categories~\cite{zhang2021discovering,zhang-2022-new-intent-discovery,zhou2023latent,usnid,DWGF}.
\begin{figure}[t]   
\centering
\includegraphics[width=1.0\columnwidth]{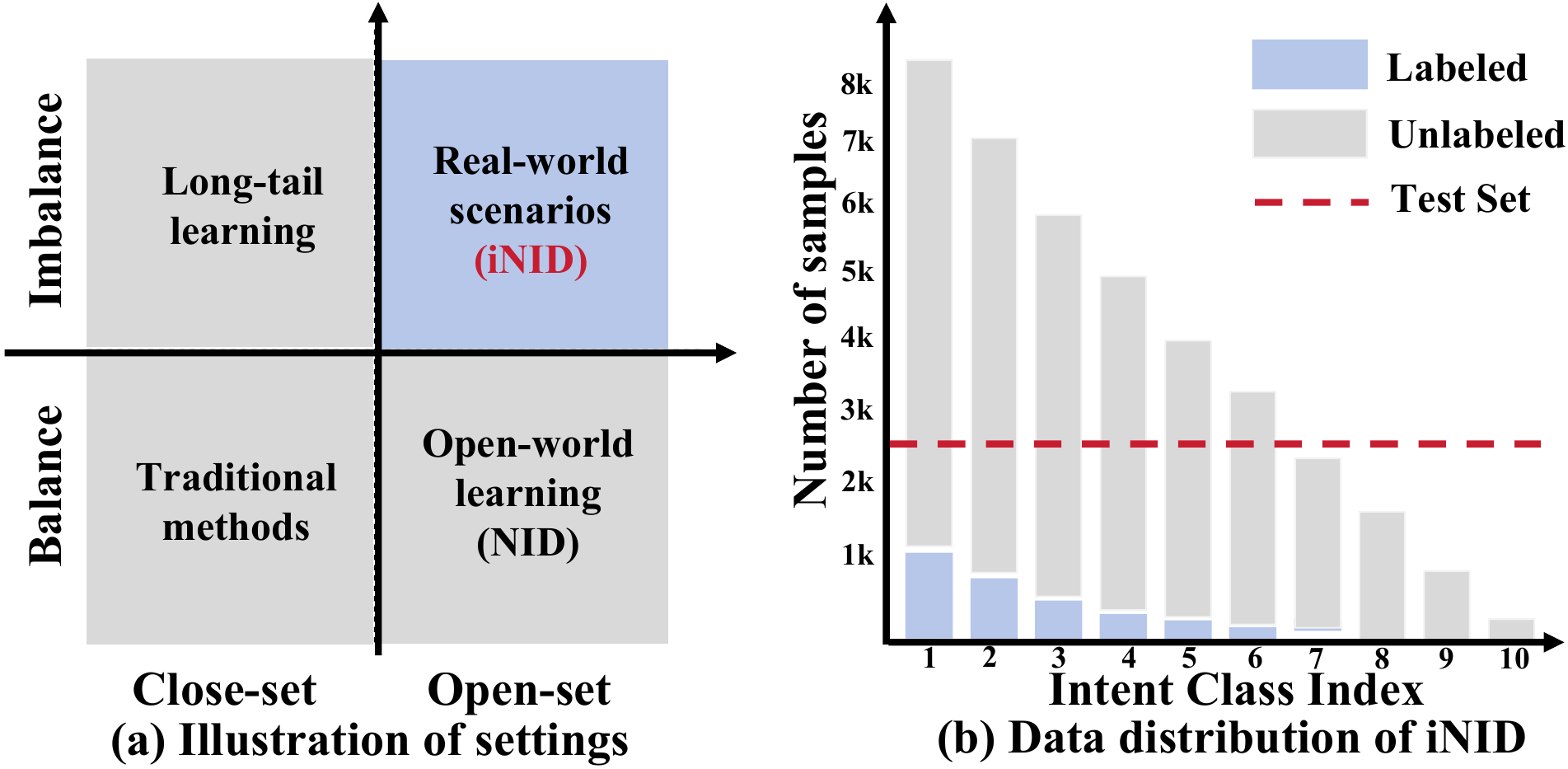}
\caption{
Illustration of proposed \taskname{} task: (a) \taskname{} unifies open-world and long-tail learning paradigms; (b) \taskname{} uses labeled and unlabeled data following a long-tail distribution to identify and categorize user intents.
\vspace{-10pt}
}
\label{fig:intro_}
\end{figure}

%

Despite the considerable advancements in NID, there remain two salient challenges impeding adoption in practical scenarios. In Fig.~\ref{fig:intro_}, most NID approaches predominantly address the issue of intent discovery within the framework of balanced datasets. But the distribution of intents often follows a long-tailed pattern~\cite{gid}, particularly in dialogue systems, wherein a small number of intents are highly represented and a wide variety of intents (unknown intents) are sparsely exemplified. Secondly, NID methods suffer from severe clustering degradation, where lack of improved methods for unbalanced data distributions and leading to poor performance in unbalanced scenarios. Therefore, we explore the new methods under the \textbf{I}mbalanced \textbf{N}ew \textbf{I}ntent \textbf{D}iscovery (\textbf{\taskname{}}) task to bridge the gap between the NID and real-world applications.

%
%

%
%
To break out the aforementioned limitations, we propose a novel framework \modelname{}, which includes three key components: model pre-training, reliable pseudo-labeling (RPL), and robust representation learning (RRL).
Specifically, the multi-task pre-training incorporates the generalized prior knowledge into the mode for establishing a robust representational foundation conducive to clustering known and novel intents.
The RPL component formulates the pseudo-label generation as a relaxed optimal transport problem, applying adaptive constraints to recalibrate the class distribution for enhanced uniformity. The model bias issues can be mitigated in long-tail settings while furnishing reliable supervisory cues for downstream representation learning.
Then, a novel distribution-aware and quality-aware noise regularization technique is introduced in RRL to effectively distinguish between clean and noisy samples. A contrastive loss function is subsequently used to facilitate the formation of distinct and well-separated clusters of representations for known and novel intent categories.
The collaborative synergy between RPL and RRL fosters an iterative training process to create a symbiotic relationship. This iterative approach cultivates intent representations conducive to clustering, significantly aiding the \taskname{} task. For better evaluation of unbalanced distribution, we introduce a comprehensive benchmark \dataset{} for \taskname{} evaluation.

Extensive experiments of \modelname{} are evaluated on the previous common benchmarks and our proposed benchmark \dataset{}. The results demonstrate that \modelname{} consistently achieves state-of-the-art performance across all clusters, notably surpassing standard NID models by an average margin of 2.7\% in long-tailed scenarios. The contributions are summarized as follows:
\begin{itemize}
\item We introduce the imbalanced new intent discovery (\taskname{}) task, which first encapsulates the challenges of clustering known and novel intent classes within long-tailed distributions. Different model performances under unbalanced distribution are sufficiently explored.

\item We construct three comprehensive \taskname{} datasets to facilitate further advancements in \taskname{} research. Our extensive experiments on these datasets validate the superiority of the proposed method \modelname{}.

\item For \taskname{}, we develop a novel \modelname{} approach that iteratively enhances pseudo-label generation and representation learning to ensure cluster-adaptive intent representations.

\end{itemize}

\section{Datasets}
We introduce a new benchmark (called \dataset{}) for NID evaluation tailored to long-tail distribution scenarios, which comprises three datasets named CLINC150-LT, BANKING77-LT, and StackOverflow20-LT, derived from CLINC~\cite{larson2019clinc}, BANKING~\cite{casanueva2020efficient} and StackOverflow~\cite{xu2015short}. Comprehensive statistics for each dataset are documented in Appendix~\ref{sec:appendix_Datasets}. Here, we describe the details of the \dataset{} datasets.
\begin{table}[]
\resizebox{0.95\columnwidth}{!}{
    \begin{tabular}{l|c|c|c|c|c}
    \toprule
    \dataset{} & $|\mathcal{Y}^{k}|$  & $|\mathcal{Y}^{n}|$ & $|\mathcal{D}_{l}|$ & $|\mathcal{D}_{u}|$ & $|\mathcal{D}_{t}|$ \\
    \midrule
     CLINC150-LT & 113 & 37 & 583 & 6395 & 2250  \\
     BANKING77-LT & 58 & 19 & 383 & 4658 & 3080  \\
     StackOverflow20-LT & 15 & 5 & 510 & 6669 & 1000  \\
    \bottomrule
    \end{tabular}}
    \caption{Statistics of the~\dataset{} datasets when $\gamma=10$. 
    $|\mathcal{Y}^{k}|$, $|\mathcal{Y}^{n}|$,  $|\mathcal{D}_{l}|$,  $|\mathcal{D}_{u}|$ and $|\mathcal{D}_{t}|$ represent the number of known categories, novel categories, labeled data, unlabeled data, and testing data.}
    \label{tab:stastic_datasets}
    \vspace{-5pt}
\end{table}    

\begin{figure}[t]
    \centering
    \includegraphics[width=0.7\linewidth]{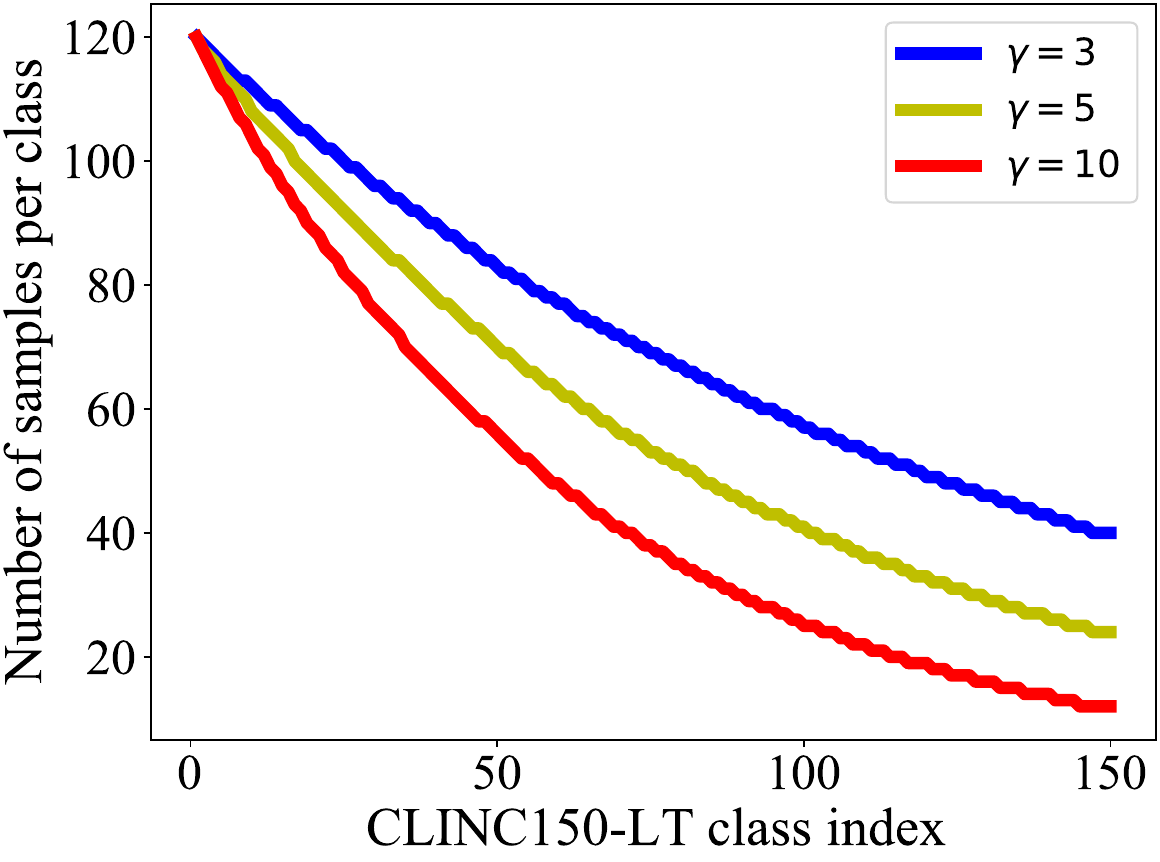}
    \caption{Number of training samples per class in artificially created long-tailed CLINC150-LT datasets with different imbalance factors.}
    \label{fig:Imbalanced_clinc}
    \vspace{-15pt}
\end{figure}

\begin{figure*}[t]
    \centering
    \includegraphics[width=1.0\linewidth]{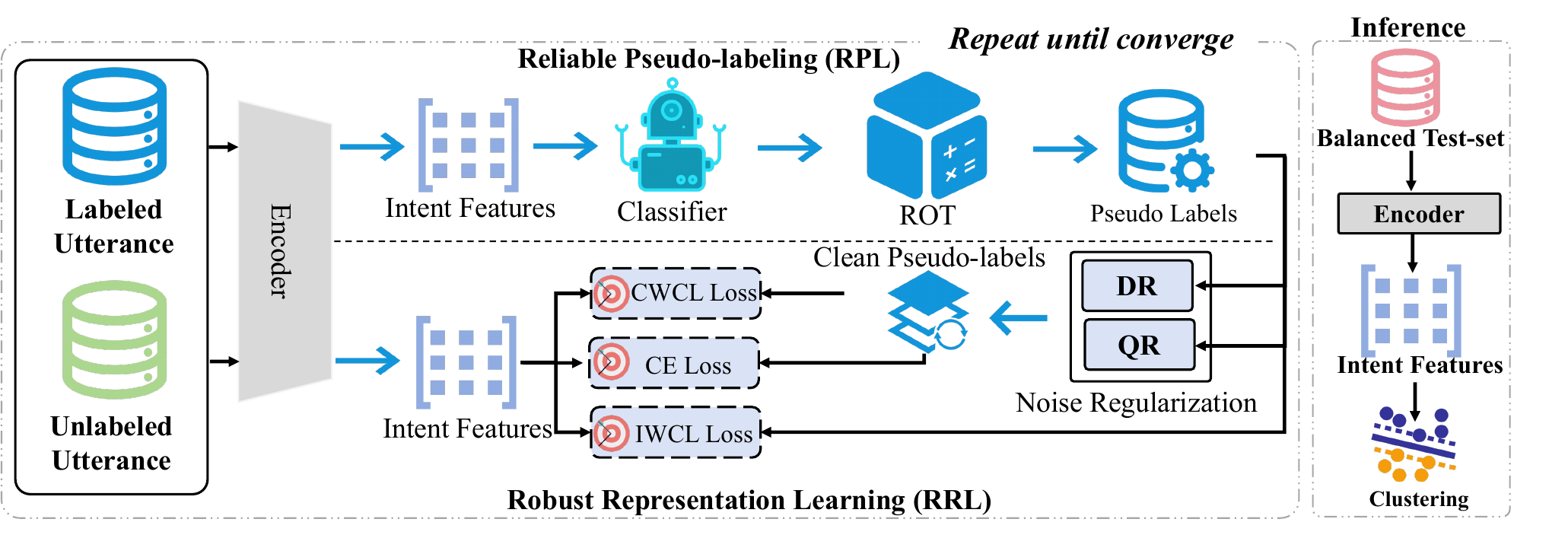}
    \caption{
    Overview of \modelname{}. The relaxed optimal transport (\texttt{ROT}) technique is used to produce high-quality pseudo-labels.
    Distribution-aware regularization (\texttt{DR}) and quality-aware regularization (\texttt{QR}) aim at filtering clean pseudo-labels. Finally, our framework incorporates class-wise contrastive learning (\texttt{CWCL}) and instance-wise contrastive learning (\texttt{IWCL}) to embed the data into a representation space where similar samples cluster together.
    }
    \label{fig:model}
\end{figure*}

\paragraph{Data Construction}
%
The first step is to simulate the long-tail distribution frequently encountered in real-world scenarios~\cite{imbalance_gama}. Each class is assigned an index $i$ ($1 \le k \le K$), where $K$ denotes the total number of intent categories.
$\gamma=\frac{n_{max}}{n_{min}}$ denotes the imbalance ratio, where $n_k$ denotes the data size of class $k$, $n_{max}=\max_{1 \leq k \leq K}({n_k})$, and $n_{min}=\min_{1 \leq k \leq K}({n_k})$. We sample from each class based on $n_{k}=n_{\max} \gamma^{(j-1) / K}$.
To explore the impact of data imbalance in NID, we construct \dataset{} by sampling with diverse imbalance ratios $\gamma \in \{3, 5, 10\}$. Fig.~\ref{fig:Imbalanced_clinc} shows the datasets created for CLINC150-LT with different imbalance factors (More details can be found in Appendix~\ref{sec:appendix_Datasets}). To simulate an open-world NID setting. We randomly select $75\%$ of intents as known intents, and sample only $10\%$ instances from known intent categories to form a labeled subset, while the remaining instances are treated as unlabeled data.
\paragraph{Data Statistics}
Since different proportions of imbalance ratios $\gamma$ have different statistics, here we only display the results of $\gamma=10$ for brevity. 
Table~\ref{tab:stastic_datasets} shows the statistics of CLINC150-LT, BANKING77-LT and StackOverflow20-LT.
We will release these datasets for future research.
\section{Methodology}
\subsection{\taskname{}} 
Supposing we have a set of labeled intent data $\mathcal{D}_{l} = \{(x_{i},y_{i})|y_{i} \in \mathcal{Y}^{k}\}$ only comprised of known intent categories $\mathcal{Y}^{k}$, the deployed model in the wild may encounter inputs from unlabeled data $\mathcal{D}_{u} = \{x_{i}|y_{i} \in \{\mathcal{Y}^{k}, \mathcal{Y}^{n} \}\}$.The unlabeled data $\mathcal{D}_{u}$ contains both known intent categories $\mathcal{Y}^{k}$ and novel intent categories $\mathcal{Y}^{n}$, where $\mathcal{Y}^{k}$ and $\mathcal{Y}^{n}$ denote the data with the \textbf{K}nown and \textbf{N}ovel intents data, respectively.
Both $\mathcal{D}_{l}$ and $\mathcal{D}_{u}$ present a long-tail distribution with imbalance ratio $\gamma > 1$.
The goal of~\taskname{} is to classify known classes and cluster novel intent classes in $\mathcal{D}_{u}$ by leveraging $\mathcal{D}_{l}$.
Finally, model performance will be evaluated on a balanced 
testing set $\mathcal{D}_{t} = \{(x_{i},y_{i})|y_{i} \in \{\mathcal{Y}^{k}, \mathcal{Y}^{n} \}\}$.

\subsection{Overall Framework} 
To achieve the learning objective of~\taskname{}, we propose an iterative method to bootstrap model performance on reliable pseudo-labeling and robust representation learning.
As shown in Fig.~\ref{fig:model}, our model mainly consists of three stages.
Firstly, we pre-train a feature extractor on both labeled and unlabeled data to optimize better knowledge transfer (Sec.~\ref{sec:Model Pre-training}).
Secondly, we obtain more accurate pseudo-labels by solving a relaxed optimal transport problem (Sec.~\ref{sec:ROT}).
Thirdly, we propose two noise regularization techniques to divide pseudo-labels and employ contrastive loss to generate well-separated clusters of representations for both known and novel intent categories(Sec.~\ref{sec:Robust Representation Learning}).

\subsection{Model Pre-training}
\label{sec:Model Pre-training}

\paragraph{Intent Representation Extraction}
To trigger the power of pre-trained language models in NID, we use BERT \cite{Devlin2019BERTPO,GanLM} as the intent encoder ($E_{\theta}:\mathcal{X} \rightarrow \mathbb{R}^{H})$. 
Firstly, we feed the $i^{th}$ input sentence $x_{i}$ to BERT, 
and take all token embeddings $[t_0, \dots, t_M]$ $\in$ $\mathds R^{(M+1) \times H}$ 
from the last hidden layer ($t_0$ is the embedding of the \texttt{[CLS]} token). 
The mean pooling is applied to get the averaged sentence representation $\boldsymbol{z}_{i} \in \mathds R^{H}$:
\begin{BigEquation}
\begin{align}
    \boldsymbol{z}_{i} = \frac{1}{M + 1}\sum_{i=0}^{M}t_{i}
\end{align}
\end{BigEquation}where $\texttt{[CLS]}$ is the vector for text classification, $M$ is the sequence length, and $H$ is the hidden size. 

\paragraph{Knowledge Sharing} 
To effectively generalize prior knowledge through pre-training to unlabeled data, we fine-tuned BERT on labeled data ($\mathcal{D}_{l}$) using the cross-entropy (CE) loss and on all available data ($\mathcal{D}_{a} = \mathcal{D}_{l} \cup \mathcal{D}_{u}$) using the masked language modeling (MLM) loss. The training objective of the fine-tuning can be formulated as follows:
\begin{BigEquation}
\begin{equation}
     \mathcal{L}_{p} = -\mathbb{E}_{x \in \mathcal{D}_{l}} \log P(y|x) - \mathbb{E}_{x \in \mathcal{D}_{a}} \log P(\hat{x}|x_{\backslash m(x)})
\end{equation}
\end{BigEquation}where $\mathcal{D}_{l}$ and $\mathcal{D}_{u}$ are labeled and unlabeled intent corpus, respectively. $P(\hat{x}|x_{\backslash m(x)}) $predicts masked tokens $\hat{x}$ based on the masked sentence $x_{\backslash m(x)}$, where $m(x)$ denotes the masked tokens. The model is trained on the whole corpus $\mathcal{D}_{a} = \mathcal{D}_{l} \cup \mathcal{D}_{u}$.

\subsection{Reliable Pseudo-labeling}
\label{sec:ROT}
\paragraph{Optimal Transport}
Here we briefly recap the well-known formulation of optimal transport (OT). Given two probability simplex vectors  $\boldsymbol{\alpha}$  and  $\boldsymbol{\beta}$  indicating two distributions, as well as a cost matrix $\mathbf{C} \in \mathbb{R}^{|\boldsymbol{\alpha}| \times|\boldsymbol{\beta}|}$ , where  $|\boldsymbol{\alpha}|$ denotes the dimension of  $\boldsymbol{\alpha}$, OT aims to seek the optimal coupling matrix $\mathbf{Q}$  by minimizing the following objective:
\begin{BigEquation}
\begin{equation}
\label{eq:OT}
\min _{\mathbf{Q} \in \boldsymbol{\Pi}(\boldsymbol{\alpha}, \boldsymbol{\beta})}\langle\mathbf{Q},\mathbf{C}\rangle
\end{equation}
\end{BigEquation}where $\langle\cdot, \cdot\rangle$ denotes frobenius dot-product. 
The coupling matrix $\mathbf{Q}$ satisfies the polytope  $\boldsymbol{\Pi}(\boldsymbol{\alpha}, \boldsymbol{\beta})= \left\{\mathbf{Q} \in \mathbb{R}_{+}^{|\bm{\alpha}| \times|\boldsymbol{\beta}|} \mid \mathbf{Q} \bm{1}_{|\boldsymbol{\beta}|}=\boldsymbol{\alpha}, \mathbf{Q}^{\top}\bm{1}_{|\boldsymbol{\alpha}|}=\boldsymbol{\beta}\right\}$, where $\boldsymbol{\alpha}$ and $\boldsymbol{\beta}$ are essentially marginal probability vectors. 
Intuitively speaking, these two marginal probability vectors can be interpreted as coupling budgets, which control the mapping intensity of each row and column in $\mathbf{Q}$.
\paragraph{Relaxed Optimal Transport for Pseudo-labeling}
The variables $\textbf{Q}\in\mathbb{R}_{+}^{N \times K}$ and $\textbf{P}\in\mathbb{R}_{+}^{N \times K}$ represent pseudo-labels matrix and classifier predictions, respectively,
where $N$ is the number of samples, and $K$ \footnote{We estimate the number of classes $K$ based on previous works~\cite{zhang2021discovering} to ensure a fair comparison. 
We provide a detailed discussion on estimating $K$ in Appendix~\ref{sex:appidx_estimate_K}.} is the number of classes.
The OT-based PL considers mapping samples to class and the cost matrix $\mathbf{C}$ can be formulated as $-\log\mathbf{P}$. 
So, we can rewrite the objective for OT-based PL based on the problem (\ref{eq:OT}) as follows:
\begin{BigEquation}
\begin{equation}
\label{eq:OT_H}
\begin{aligned}
&\min_{\mathbf{Q},\bm{b}}{\langle \mathbf{Q}, {-\log{\mathbf{P}}} \rangle} + \lambda H(\mathbf{Q}) \\
&s.t.\,\,\mathbf{Q}\bm{1}=\bm{\alpha},\mathbf{Q}^T\bm{1}=\bm{\beta}, \mathbf{Q}\geq0
\end{aligned}
\end{equation}
\end{BigEquation}where the function $H$ is the entropy regularization, $\lambda$ is a scalar factor, $\bm{\alpha}=\frac{1}{N}\bm{1}$ is the sample distribution and $\bm{\beta}$ is class distribution. So the pseudo-labels matrix $\mathcal{U}_{a}$ can be obtained by normalization: $N\mathbf{Q}$.
However, in the~\taskname{} setup, the class distribution is often long-tailed and unknown, and the model optimized based on the problem (\ref{eq:OT_H}) tends to learn degenerate solutions. This mismatched class distribution will lead to unreliable pseudo-labels.
To mitigate this issue, we impose a soft constraint (\texttt{ROT}) on the problem (\ref{eq:OT_H}).
Instead of the traditional equality constraint~\cite{naive_OT,OT_2020_Mathilde_Caron}, 
we employ a Kullback-Leibler (KL) divergence constraint to encourage a uniform class distribution. 
This constraint is crucial for preventing degenerate solutions in long-tailed scenarios while allowing for the generation of imbalanced pseudo-labels due to its more relaxed nature compared to an equality constraint.
%
The formulation of ROT is articulated as follows:
\begin{BigEquation}
\begin{equation}
\begin{aligned}
&\min_{\mathbf{Q},\bm{\beta}}{\langle \mathbf{Q}, {-\log{\mathbf{P}}} \rangle} + \lambda_1 H(\mathbf{Q}) + \lambda_2 D_{\text{KL}}(\frac{1}{K}\bm{1}, \bm{\beta}) \\
&s.t.\,\,\mathbf{Q}\bm{1}=\bm{\alpha},\mathbf{Q}^T\bm{1}=\bm{\beta}, \mathbf{Q}\geq0, \bm{\beta}^T\bm{1}=1
\end{aligned}
\label{eq:OT_KL}
\end{equation}
\end{BigEquation}where $\lambda_2$ is a hyper-parameter and $D_{\text{KL}}$ is the Kullback-Leibler divergence.
The optimization problem (\ref{eq:OT_KL}) can be tractably solved using the Sinkhorn-Knopp algorithm~\cite{{cuturi2013sinkhorn}} and we detail the optimization process in Appendix~\ref{sec:ROT_optimal}.

\subsection{Robust Representation Learning}
\label{sec:Robust Representation Learning}
Directly using generated pseudo-labels for representational learning is risky due to significant noise in early-stage pseudo-labeling.
Consequently, we categorize pseudo-labels as clean or noisy based on their distribution and quality, applying contrastive loss to achieve cluster-friendly representations.

\noindent
\noindent
\textbf{Noise Regularization} 
We initially introduce a \textit{distribution-aware regularization} (\texttt{DR}) to align the sample selection ratio with the class prior distribution, effectively mitigating selection bias in~\taskname{} setup.
This regularization combines small-loss instances with class distributions, ensuring inclusive representation of all classes, particularly \texttt{Tail} categories, during training.
Specifically, the final set of selected samples $S^\prime$ is represented as follows:
\begin{BigEquation}
\begin{equation}
S^\prime = \bigcup_{j=1}^{K} s_{j}^{\prime}
\end{equation}
\end{BigEquation}where $K$ is total classes, $s_{j}^{\prime}$ is the set of samples selected from the $j$-th category slice $s_j$, defined as:
\begin{equation}
s_{j}^{\prime} = \left\{ h \mid (h \in s_{j}) \wedge (\text{sort}(l(h)) \leq k_j) \right\}
\label{eq:distribution-aware}
\end{equation}
where $l(h)$ is the instance-level loss of $h$,  $\rho$ is threshold hyper-parameter, $r_{j}$ is the class distribution, $k_j=\min(\left|s_{j}\right|, \lceil N \rho r_{j} \rceil)$.

In addition, to select high-confidence pseudo-labels that closely align with the predicted labels, we propose a \textit{quality-aware regularization} (\texttt{QR}).
Specifically, we calculate confidence scores for each pseudo-label and then select the clean samples, denoted as $h$, whose confidence scores exceed a certain threshold $\tau_{g}$:
\begin{BigEquation}
\begin{equation}
\mathcal{A}^\prime=\left\{h \mid\left(h \in \mathcal{U}_{a
}\right) \wedge \left(\operatorname{max}\left(\bm{p}\right) > \tau_{g} \right)\right\}
\label{eq:quality-aware}
\end{equation}
\end{BigEquation}where $\bm{p}$ is the probability vector for $h$ and $\tau_{g} \in [0,1]$ is a confidence threshold hyper-parameter.
Then the overall pseudo-labels $\mathcal{U}_{a}$ can filter out the clean pseudo-labels $\mathcal{U}_{clean}$ as follows:
\begin{BigEquation}
\begin{equation}
\begin{aligned}
\mathcal{U}_{clean} &=\left\{h \mid\left(h \in \mathcal{S}^\prime\right) \lor \left(h \in \mathcal{A}^\prime\right)  \right\}
\end{aligned}
\end{equation}
\end{BigEquation}

\noindent
\textbf{Contrastive Clustering}
Following the extraction of clean pseudo-labels, we extend the traditional contrastive loss~\cite{SCL2020} to utilize label information, forming positive pairs from same-class samples within $\mathcal{U}_{clean}$.
Additionally, to enhance the model's emphasis on clean samples, we introduce a method for encoding soft positive correlation among pseudo-positive pairs, enabling adaptive contribution. Specifically, for an intent sample $x_i$, we first acquire its $L2$-normalized embedding $z_i$. By multiplying the confidence scores $q$ of two samples, we obtain an \textit{adaptive weight} $w_{ij}=q_i \cdot q_j$. The class-wise contrastive loss (\texttt{CWCL}) is then defined as follows:
\begin{BigEquation}
\begin{align}
\begin{split}
    \mathcal{L}_{c}(i)&=\sum_{p \in P(i)} w_{i p} \cdot \log \frac{\exp \left(z_{i} \cdot z_{p} / \tau\right)}{\sum_{j} \mathbbm{1}_{i \neq j} \exp \left(z_{i} \cdot z_{j} / \tau\right)} \\
    P(i)&=\left\{p \mid \left(p \in \mathcal{U}_{clean}\right) \wedge \left(c_{i}=c_{p}\right)\right\}
    \label{eq:class_wise_CL}
\end{split}
\end{align}
\end{BigEquation}where $P(i)$ represents the indices of instances sharing the same label as $x_i$, and $\tau$ is a hyper-parameter. 
Fundamentally, \texttt{CWCL} loss brings intents of the same class closer together while distancing clusters of different classes, effectively creating a clustering effect.
To enhance the generalization of intent representation, we incorporate instance-wise contrastive learning~\cite{selfSCL}. 
The augmented views of instances in $\mathcal{U}_{a}$ are used as positive examples.
The instance-wise contrastive loss (\texttt{IWCL}) is defined as follows:
\begin{BigEquation}
\begin{equation}
	\begin{split}
		\mathcal{L}_{i}(i)= -\log \frac{ \exp \left(z_i \cdot \bar z_{i} / \tau\right)}{\sum_{j} \mathbbm{1}_{i \neq j} \exp \left(z_i \cdot z_j / \tau\right)}
	\end{split}
	\label{eq:instance-wise_CL}
\end{equation}
\end{BigEquation}
where $z_i$, $\bar z_{i}$ regard an anchor and its augmented sample, respectively, and $\bar z_i$ denotes the random token replacement augmented view of $z_i$.

\paragraph{Joint Training} 
To mitigate the risk of catastrophic forgetting of knowledge, we incorporate cross-entropy loss on $\mathcal{U}_{clean}$ into the training process.
Overall, the optimization of \modelname{} is to minimize the combined training objective:
\begin{BigEquation}
\begin{equation}
\mathcal{L}_{all}=\omega \cdot (\sum_{i \in N}\frac{1}{1+|P(i)|} (\mathcal{L}_{c}(i) +  \mathcal{L}_{i}(i))) + (1-\omega) \cdot \mathcal{L}_{ce}
\label{eq:final_loss}
\end{equation}
\end{BigEquation}where $\omega$ is a hyper-parameter and $|\cdot|$ is the cardinality computation. When $x_i$ is a noisy example,  $\mathcal{L}_{c}(i)=0$ and $|P(i)|=0$.
During inference, we only utilize the cluster-level head and compute the argmax to get the cluster results.

%
\begin{table*}[!t]
    \centering
    \resizebox{0.78\textwidth}{!}{
    \begin{tabular}{p{2cm}<{\centering}|c c c c c c c c c c c c}
    \toprule
        \multirow{4}{*}{\textbf{Methods}} & \multicolumn{12}{c}{\textbf{CLINC150-LT}}  \\
        \cmidrule(lr){2-13}
        & \multicolumn{3}{c}{$\gamma=1$} & \multicolumn{3}{c}{$\gamma=3$} & \multicolumn{3}{c}{$\gamma=5$} & \multicolumn{3}{c}{$\gamma=10$} \\
        \cmidrule(lr){2-4}\cmidrule(lr){5-7}\cmidrule(lr){8-10}\cmidrule(lr){11-13}
        & NMI & ARI & ACC & NMI & ARI & ACC & NMI & ARI & ACC & NMI & ARI & ACC \\ \cmidrule(lr){1-1}\cmidrule(lr){2-4}\cmidrule(lr){5-7}\cmidrule(lr){8-10}\cmidrule(lr){11-13}
        GCD & 91.13 & 67.44 & 77.50 & 87.61 & 59.71 & 73.07 & 84.18 & 53.04 & 67.96 & 80.21 & 47.64 & 61.91\\
        DeepAligned & 93.89 & 79.75 & 86.49 & 92.29 & 73.79 & 81.78 & 90.93 & 70.19 & 79.02 & 88.43 & 62.47 & 71.47\\
        CLNN & 95.45 & 84.30 & 89.46 & 93.52 & 78.02 & 85.42 & 92.54 & 73.05 & 79.38 & 89.52 & 63.92 & 72.00 \\
        DPN & 95.11 & 86.72 & 89.06 & 94.84 & 79.98 & 85.64 & 94.51 & 79.32 & 84.49 & 92.43 & 70.62 & 77.51\\
        LatentEM & 95.01 & 83.00 & 88.99 & 93.74 & 78.16 & 84.62 & 93.39 & 77.23 & 83.78 & 92.01 & 72.77 & 80.22\\
        USNID & 96.55 & 88.43 & 92.18 & 94.67 & 80.30 & 85.33 & 94.06 & 77.60 & 82.49 & 91.62 & 68.61 & 74.40 \\
        \rowcolor{Gray} \modelname{} & \textbf{97.26} & \textbf{91.78} & \textbf{95.64} & \textbf{95.60} & \textbf{85.36} & \textbf{90.44}& \textbf{94.65} & \textbf{81.90} &  \textbf{88.04} & \textbf{93.40} & \textbf{76.21}& \textbf{82.40}\\
        \bottomrule[0.5pt]
    \toprule[0.5pt]
    
       \multirow{4}{*}{\textbf{Methods}} & \multicolumn{12}{c}{\textbf{BANKING77-LT}}  \\
        \cmidrule(lr){2-13}
        & \multicolumn{3}{c}{$\gamma=1$} & \multicolumn{3}{c}{$\gamma=3$}& \multicolumn{3}{c}{$\gamma=5$}& \multicolumn{3}{c}{$\gamma=10$} \\
        \cmidrule(lr){2-4}\cmidrule(lr){5-7}\cmidrule(lr){8-10}\cmidrule(lr){11-13}
        & NMI & ARI & ACC & NMI & ARI & ACC & NMI & ARI & ACC& NMI & ARI & ACC \\ \cmidrule(lr){1-1}\cmidrule(lr){2-4}\cmidrule(lr){5-7}\cmidrule(lr){8-10}\cmidrule(lr){11-13}
        GCD & 77.86 & 46.87 & 58.95 & 71.92 & 42.35 & 56.98 & 69.16 & 37.93 & 53.41 & 66.89 & 33.38 & 46.92\\
        DeepAligned & 79.39 & 53.09 & 64.63 & 78.93 & 51.65 & 63.64 & 77.99 & 48.56 & 60.06 & 75.01 & 44.11 & 54.03\\
        CLNN & 86.19 & 66.98 & 77.22 & 85.64 & 65.34 & 75.75 & 82.95 & 58.87 & 70.65 & 79.99 & 52.04 & 62.63 \\
        DPN & 82.58 & 61.21 & 72.96 & 84.43 & 61.36 & 72.27 & 80.88 & 49.75 & 61.69 & 77.17 & 43.41 & 57.95\\
        LatentEM & 84.02 & 62.92 & 74.03 & 83.37 & 61.23 & 73.08 & 81.38 & 56.78 & 69.51 & 80.55 & 55.65 & 65.05\\
        USNID & 87.53 & 69.88 & 79.92 & 86.62 & 67.01 & 75.03 & 83.59 & 60.56 & 70.06 & 80.49 & 54.26 & 63.15 \\
        \rowcolor{Gray} \modelname{} & \textbf{87.66} & \textbf{70.13} & \textbf{81.14} & \textbf{86.79} & \textbf{67.35} & \textbf{76.72} & \textbf{83.60} & \textbf{61.18} & \textbf{72.89}& \textbf{81.08} & \textbf{55.80} & \textbf{66.59}\\
    \bottomrule[0.5pt]
    \toprule[0.5pt]
       \multirow{4}{*}{\textbf{Methods}} & \multicolumn{12}{c}{\textbf{StackOverflow20-LT}}  \\
        \cmidrule(lr){2-13}
        & \multicolumn{3}{c}{$\gamma=1$} & \multicolumn{3}{c}{$\gamma=3$}& \multicolumn{3}{c}{$\gamma=5$} & \multicolumn{3}{c}{$\gamma=10$} \\
        \cmidrule(lr){2-4}\cmidrule(lr){5-7}\cmidrule(lr){8-10}\cmidrule(lr){11-13}
        & NMI & ARI & ACC & NMI & ARI & ACC & NMI & ARI & ACC& NMI & ARI & ACC \\ \cmidrule(lr){1-1}\cmidrule(lr){2-4}\cmidrule(lr){5-7}\cmidrule(lr){8-10}\cmidrule(lr){11-13}
        GCD & 62.07 & 45.11 & 66.81 & 61.86 & 40.59 & 65.30 & 57.84 & 36.15 & 59.10 & 48.04 & 27.55 & 48.60\\
        DeepAligned & 76.47 & 62.52 & 80.26 & 75.27 & 62.73 & 77.10 & 75.47 & 64.19 & 78.50 & 73.47 & 61.82 & 73.80\\
        CLNN & 77.12 & 69.36 & 82.90 & 78.78 & 68.98 & 84.30 & 77.67 & 65.81 & 76.70 & 75.29 & 60.46 & 76.60 \\
        DPN & 61.13 & 52.59 & 48.09 & 79.64 & 69.22 & 85.00 & 78.91 & 51.81 & 81.00 & 76.56 & 63.15 & 78.30\\
        LatentEM & 77.32 & 65.70 & 80.50 & 75.54 & 63.04 & 77.40 & 77.42 & 65.72 & 79.20 & 77.07 & 65.20 & 78.17\\
        USNID & 81.47 & 76.08 & 86.43 & 81.99 & 74.64 & 86.90 & 81.34 & 72.28 & 83.00 & 78.09 & 66.24 & 78.90 \\
        \rowcolor{Gray} \modelname{} & \textbf{83.52} & \textbf{77.06} & \textbf{88.30} & \textbf{82.12} & \textbf{75.09} & \textbf{87.40} & \textbf{81.42} & \textbf{73.09} & \textbf{86.50}& \textbf{79.78} & \textbf{71.15} & \textbf{82.60}\\ 
    \bottomrule
    \end{tabular}
    }
    \caption{The main results on three datasets under various imbalance ratios $\gamma$ ($\gamma=1$ is the balanced NID setting). We set the known class ratio $|\mathcal{Y}^{k}|/|\mathcal{Y}^{k} \cap \mathcal{Y}^{n}|$ to 0.75, and the labeled ratio of known intent classes to 0.1 to conduct experiments. Results are averaged over three random run ($p$-value < 0.01 under t-test). We bold the \textbf{best result}.}
    \label{tab:Main_Results}
\end{table*}

\section{Experiments}
\subsection{Experimental Setup}
%

%
\paragraph{Baseline Methods} We compare our method with various baselines and state-of-the-art methods, including
DeepAligned~\cite{zhang2021discovering},
GCD~\cite{GCD},
CLNN~\cite{zhang-2022-new-intent-discovery},
DPN~\cite{DPN},
LatentEM~\cite{zhou2023latent}, and USNID~\cite{usnid}.
Please see Appendix~\ref{sec:appendix_Comparison} for more comprehensive comparison and implementation details.
\paragraph{Evaluation Metrics} 
We adopt three metrics for evaluating clustering results: Normalized Mutual Information (\texttt{NMI}), Adjusted Rand Index (\texttt{ARI}), and clustering Accuracy (\texttt{ACC}) based on the Hungarian algorithm.
Furthermore, to more easily assess the impact of long tail distribution on performance, we divide $\mathcal{Y}^{k}$ and $\mathcal{Y}^{n}$ into three distinct groups $\{$\texttt{Head}, \texttt{Medium}, \texttt{Tail}$\}$ with the proportions $|\texttt{Head}|:|\texttt{Medium}|:|\texttt{Tail}|=3:4:3$.
%
\paragraph{Implementation Details}
To ensure a fair comparison for~\modelname{} and all baselines, we adopt the pre-trained 12-layer bert-uncased BERT model\footnote{\url{https://huggingface.co/bert-base-uncased}} \cite{Devlin2019BERTPO} as the backbone encoder in all experiments and only fine-tune the last transformer layer parameters to expedite the training process \cite{zhang2021discovering}.
We adopt the AdamW optimizer with the weight decay of 0.01 and gradient clipping of 1.0 for parameter updating.
%
%
For CLNN~\cite{zhang-2022-new-intent-discovery}, the external dataset is not used as in other baselines, the parameter of top-k nearest neighbors is set to $\{$100, 50, 500$\}$ for CLINC, BANKING, and StackOverflow, respectively, as utilized in~\citet{zhang-2022-new-intent-discovery}. 
For all experiments, we set the batch size as 512 and the temperature scale as $\tau$ = 0.07 in Eq.~\eqref{eq:class_wise_CL} and Eq.~\eqref{eq:instance-wise_CL}.
We set the parameter $\rho$ = 0.7 in Eq.~\eqref{eq:distribution-aware} and the confidence threshold $\tau_{g}$ = 0.9 in Eq.~\eqref{eq:quality-aware}.
We adopt the data augmentation of random token replacement as~\citet{zhang-2022-new-intent-discovery}.
All experiments are conducted on 4 Tesla V100 GPUs and averaged over 3 runs.

\begin{figure*}[t]
    \centering
    \includegraphics[width=0.84\linewidth]{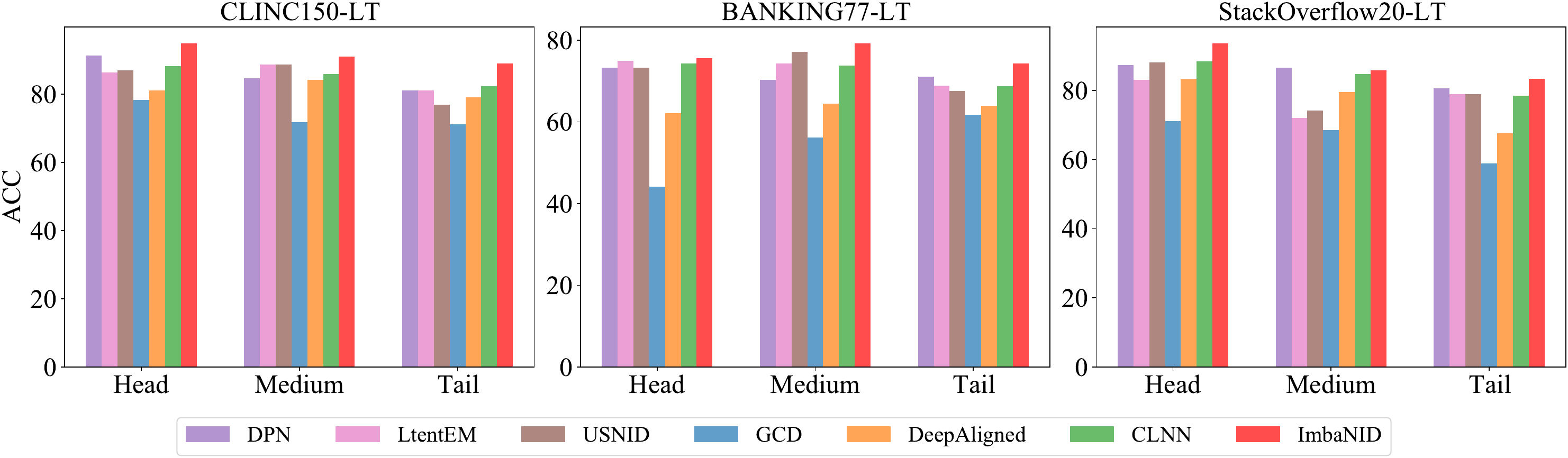}
    \caption{\texttt{Head}, \texttt{Medium}, and \texttt{Tail} comparison on the ~\dataset{} datasets.}
    \label{fig:H_M_T}
\end{figure*}
\begin{table*}[t]
    \centering
    \resizebox{0.7\textwidth}{!}{
    \begin{tabular}{l | c c c | c c c | c c c}
    \toprule
     \multirow{3}{*}{Methods} & \multicolumn{3}{c|}{CLINC150-LT} & \multicolumn{3}{c|}{BANKING77-LT} & \multicolumn{3}{c}{StackOverflow20-LT}\\
     \cmidrule{2-4} \cmidrule{5-7}  \cmidrule{8-10} 
    ~ & Head & Medium & Tail & Head & Medium & Tail & Head & Medium & Tail\\
    \midrule
     \rowcolor{Gray} \modelname{} & \textbf{82.52} & \textbf{90.67} & \textbf{71.26} & \textbf{68.26} & \textbf{66.05} & \textbf{65.87} & \textbf{90.67} & \textbf{87.25}& \textbf{81.67}\\
    \midrule
    {\large{\ding{172}}} w/ COT & 72.74 & 87.44 & 58.67 & 62.72 & 63.11 & 48.70 & 86.63 & 85.75 & 79.67 \\
    {\large{\ding{173}}} w/ EOT & 81.41 & 83.00 & 65.33 & 66.59 & 65.40 & 57.61 & 90.00 & 86.11 & 81.60 \\
    {\large{\ding{174}}} w/ MOT & 69.33 & 57.67 & 30.52 & 62.07 & 57.34 & 26.20 & 88.97 & 66.00 & 64.33 \\
    \midrule
    {\large{\ding{175}}} w/o DR & 80.74 & 88.57 & 71.21 & 67.17 & 65.08 & 49.67 & 88.33 & 86.75 & 81.33 \\
    {\large{\ding{176}}} w/o QR & 82.50 & 88.94 & 70.52 & 63.91 & 65.42 & 59.02 & 87.67 & 86.00 & 81.57 \\
    {\large{\ding{177}}} w/o DR and QR & 81.19 & 87.19 & 71.05 & 67.50 & 64.88 & 50.00 & 88.33 & 86.51 & 80.33 \\
    \midrule
    {\large{\ding{178}}} w/o Adaptive Weight & 82.37 & 90.22 & 71.11 & 68.18 & 65.81 & 64.57 & 90.30 & 87.00 & 79.67\\
    {\large{\ding{179}}} w/o CWCL & 81.93 & 90.11 & 70.81 & 67.83 & 66.03 & 58.70 &  90.33 & 85.22 & 78.00 \\
    {\large{\ding{180}}} w/o IWCL & 81.78 & 86.44 & 71.23 & 65.54 & 64.22 & 65.20 & 90.51 & 76.75 & 80.33 \\
    \bottomrule
    \end{tabular}}
    \caption{Experimental results of the ablation study on the \dataset{}~datasets at imbalance ratios $\gamma=10$.}
    \label{tab:ablation}
\end{table*}

\subsection{Main Results}
\textbf{\modelname{} achieves SOTA results in both balanced and imbalanced settings}. In Table~\ref{tab:Main_Results}, we present a comprehensive comparison of \modelname{} with prior start-of-the-art baselines in both balanced and multiple imbalanced settings.
We observe that~\modelname{} significantly outperforms prior rivals by a notable margin of 3.9\% under various settings of imbalance ratio.
Specifically,
on the broad cross-domain CLINC150-LT dataset, \modelname{} beats the previous state-of-the-art with an increase of 3.5\% in \texttt{ACC}, 0.7\% in \texttt{NMI}, and 3.9\% in \texttt{ARI} on average.
On the StackOverflow20-LT with fewer categories, \modelname{} demonstrates its effectiveness with significant improvements of 2.6\% in \texttt{ACC}, 0.6\% in \texttt{NMI}, and 2.4\% in \texttt{ARI} on average, consistently delivering substantial performance gains across each imbalanced subset.
When applied to the specific single-domain BANKING77-LT datasets, \modelname{} reliably achieves significant performance improvements, underscoring its effectiveness in narrow-domain scenarios with indistinguishable intents.
These results show the conventional NID models with naive pseudo-labeling and representation learning methods encounter a great challenge in handling the~\taskname{} task.
Our method efficiently produces accurate pseudo-labels under imbalanced conditions by employing soft constraints and utilizes these pseudo-labels to construct cluster-friendly representations.
%

\paragraph{Effectiveness on Long-tailed Distribution} We also provide a detailed analysis of the results for the \texttt{Head}, \texttt{Medium}, and \texttt{Tail} classes, offering a more comprehensive understanding of our method’s performance across three~\taskname{} datasets. Fig.~\ref{fig:H_M_T} presents the comparative accuracy among various groups under the condition $\gamma=3$. 
It is noteworthy that in \texttt{Tail} classes, the gaps between~\modelname{} and the best baseline are 4.2\%, 3.5\% and 3.7\% across three datasets. 
In contrast, most baselines exhibit degenerated performance, particularly on CLINC150-LT and BANKING77-LT.
Moreover, \modelname{} retains a competitive performance on \texttt{Head} classes. 
These results highlight the effectiveness of~\modelname{} in~\taskname{} setup, 
making it particularly advantageous for \texttt{Head} and \texttt{Tail} classes.

\subsection{Effect of Pseudo-label Assignment} 
To evaluate ROT in reliable pseudo-labels generation of the~\taskname{} setup, we compare three OT-based optimizations for pseudo-labels generation, including COT~\cite{OT_2020_Mathilde_Caron}, EOT~\cite{naive_OT}, and MOT~\cite{imbagcd}.
(1) COT denotes the removal of the \text{KL} term from our optimization problem~\eqref{eq:OT_KL}.
(2) EOT signifies the replacement of the \text{KL} term in our optimization problem~\eqref{eq:OT_KL} with a typical entropy regularization $\text{KL}(\boldsymbol{\beta}\| \hat{\boldsymbol{\beta}})$.
(3) MOT operates without any assumption on the class distribution $\boldsymbol{\beta}$, allowing $\boldsymbol{\beta}$ to be updated by the model prediction using a moving-average mechanism.
Specifically, $\boldsymbol{\beta}=\mu \hat{\boldsymbol{\beta}}+(1-\mu) \boldsymbol{v}$, where $\mu$ is the moving-average parameter, $\hat{\boldsymbol{\beta}}$ is the last updated $\boldsymbol{\beta}$ and $v_{j}=\frac{1}{N} \sum_{i=1}^{N} \mathbbm{1}\left(j=\arg \max \textbf{P}_{i}\right)$. 
From Table \ref{tab:ablation}, we can observe that \modelname{} outperforms the model \large{\ding{172}}, which indicates the necessity of imposing constraints on the class distribution.
Compared to the model \large{\ding{173}}, \modelname{} achieves the most gains for \texttt{Head} and \texttt{Tail} classes, indicating it better constrains the class distribution towards uniformity.
Finally, when compared to the above strategies, the performance of the model \large{\ding{173}} in the \texttt{Tail} classes is notably inferior. 
The results stem from inadequate constraints on the category distribution, leading to a decline in cluster quality.
The comparisons underscore that~\modelname{} demonstrates strong proficiency in generating accurate pseudo-labels within the~\taskname{} setup.

\subsection{Effect of Noise Regularization} 
To investigate the effectiveness of noise regularization (\texttt{NR}) in filtering noisy pseudo-labels, we conduct ablation experiments to analyze its contributions.
In Table~\ref{tab:ablation}, eliminating DR diminishes intent discovery performance, particularly in \texttt{Tail} classes.
This occurs because a higher proportion of \texttt{Head} classes in pseudo-labels inevitably results in model bias.
Furthermore, removing \texttt{QR} results in decreased performance, primarily because fewer examples are initially selected due to the classifier's low confidence, leading to degenerate solutions.
Notably, considering all pseudo-labels as clean leads to significant performance drops across all datasets, indicating that numerous noisy pseudo-labels may cause model overfitting and reduced generalization.
The results indicate that \texttt{NR} is indispensable to~\modelname{} in handling~\taskname{} setup.

\subsection{Effect of Contrastive Clustering} 
\begin{figure}[t]
    \centering
    \subfigure[USNID]{
    \includegraphics[width=0.45\columnwidth]{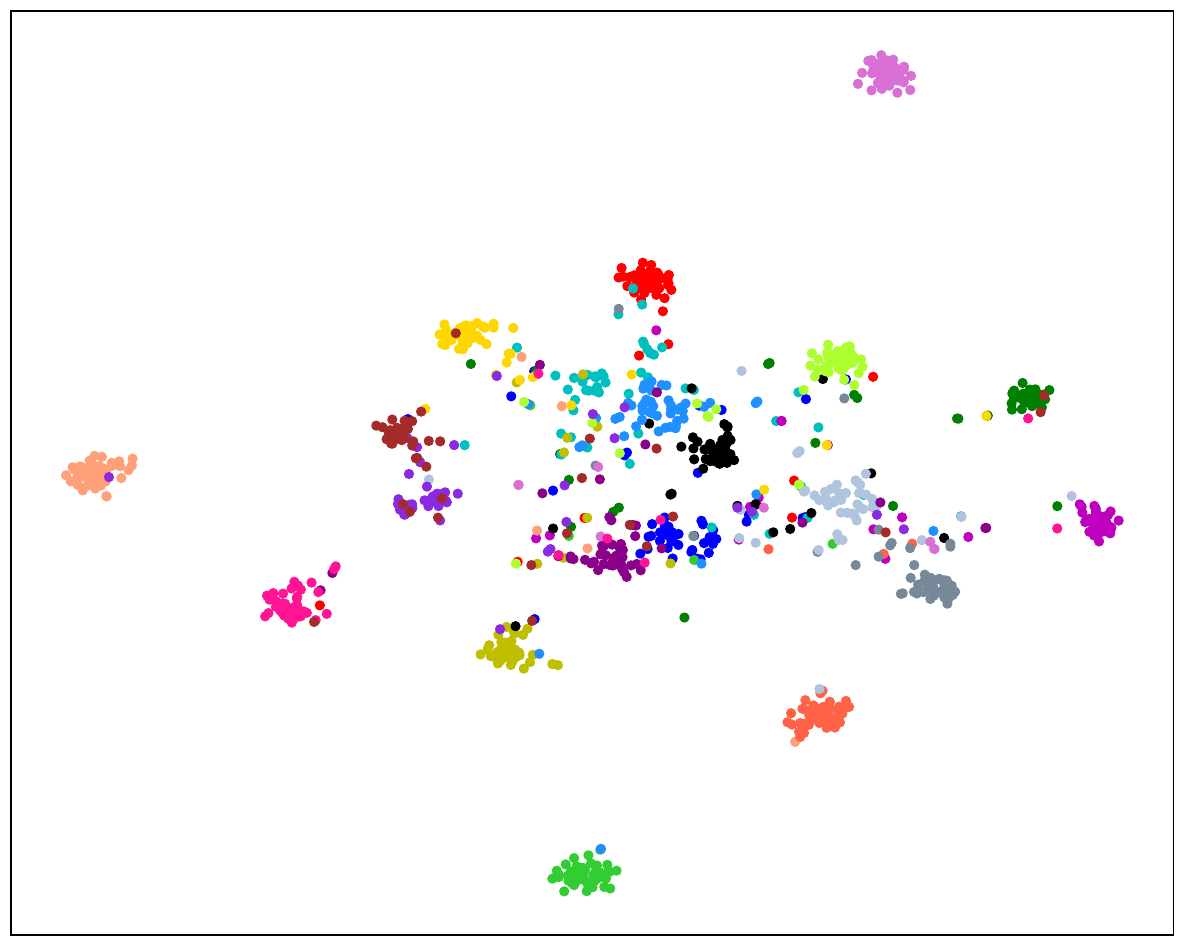}
    \label{tsne_3}
    }
    \subfigure[\modelname{}]{
    \includegraphics[width=0.45\columnwidth]{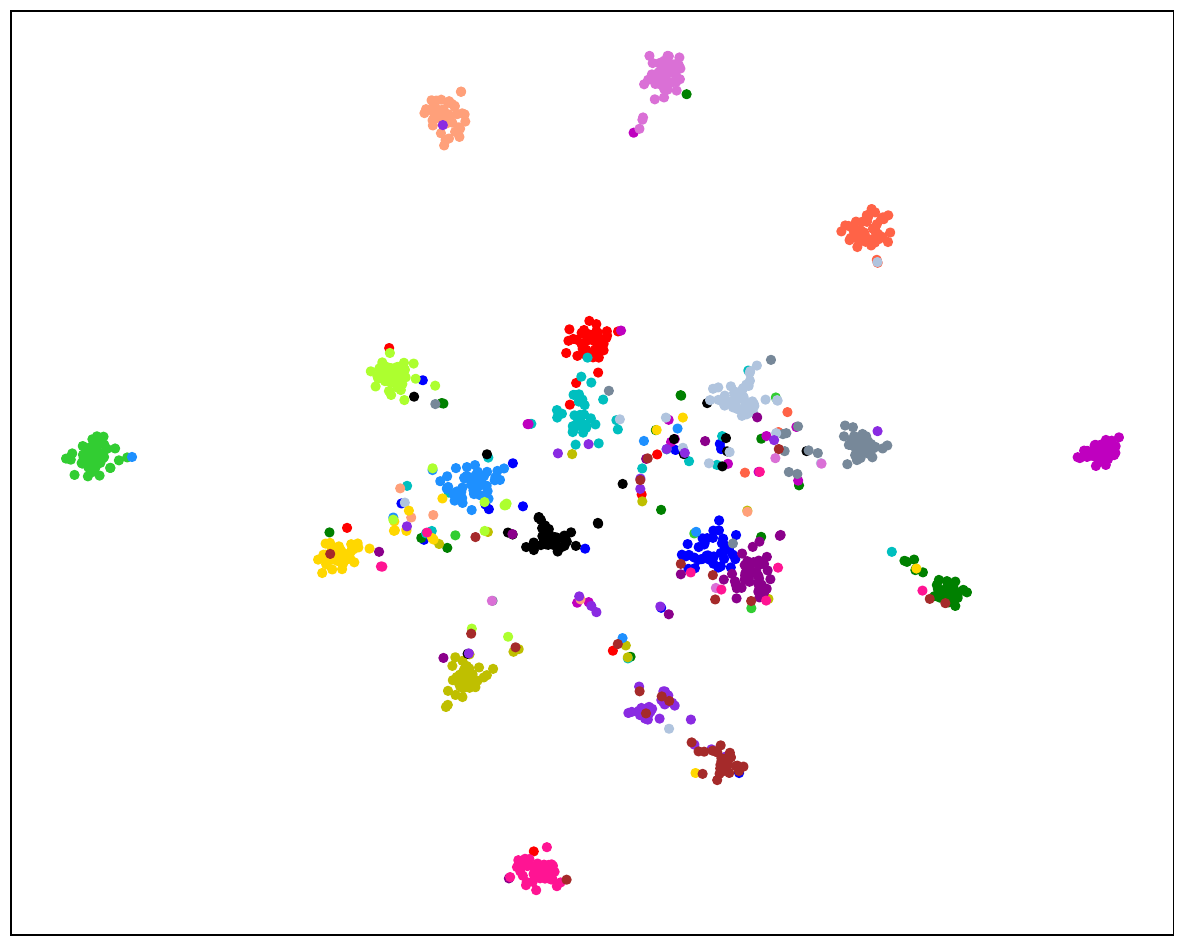}
    \label{tsne_4}
    }
    \caption{t-SNE visualization of embeddings on the StackOverflow20-LT dataset. The known class ratio $|\mathcal{Y}^{k}|/|\mathcal{Y}^{k} \cap \mathcal{Y}^{n}|$ is 0.75, and the labeled ratio is 0.1.} 
    \label{fig:t-SNE}
\end{figure}

To assess the impact of contrastive clustering in representation learning, we carry out ablation experiments to analyze its individual effects in Table~\ref{tab:ablation}.
When the adaptive weight strategy is removed from Eq.~\eqref{eq:class_wise_CL}, the model disregards probability distribution information and becomes more susceptible to noisy pseudo-labels.
Then, removing \texttt{CWCL} or \texttt{IWCL} from Eq.~\eqref{eq:final_loss} results in performance degradation, suggesting that class-wise and instance-wise contrastive learning respectively aid in developing compact cluster representations and enhancing representation generalization. 
In Fig.~\ref{fig:t-SNE}, we use t-SNE to illustrate embeddings learned on the StackOverflow20-LT dataset, where \modelname{} visibly forms more distinct clusters than comparative methods, underscoring the effectiveness of our model.

\subsection{Effect of Known Class Ratio} 
\begin{figure}[t]
    \centering
    \subfigure[Impact on CLINC-LT]{
    \includegraphics[width=0.46\columnwidth]{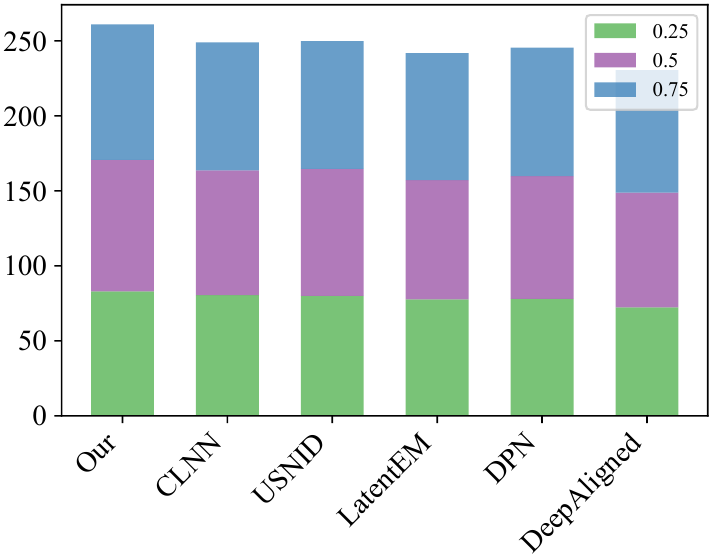}
    \label{subfigure:ratio_clinc150-LT}
    }
    \subfigure[Impact on BANKING-LT]{
    \includegraphics[width=0.46\columnwidth]{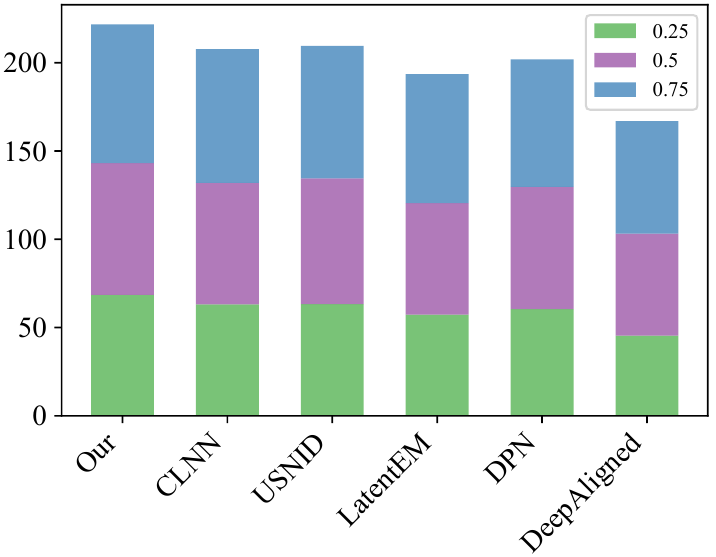}
    \label{subfigure:ratio_banking77-LT}
    }
    \caption{Impact of varying the known class ratio on two datasets. The x-axis represents different models and the y-axis denotes their corresponding accuracy values.}
    \label{fig:fig_radio_}
\end{figure}

To investigate the impact of varying numbers of known intents, we vary the ratio of known intents ranging in $\{25\%, 50\%, 75\%\}$ during training.
Fig.~\ref{fig:fig_radio_} illustrates the comparative accuracy among various ratio of known intents under the condition $\gamma=3$. We observe that even when only a few known intents are available, our method still performs better than other strong baselines.
This demonstrates its strength in learning from labeled data and discovering inherent patterns from unlabeled data.
Meanwhile, we note a rise in performance as the volume of labeled data incorporated increases, aligning with anticipated outcomes.
In short, our proposed method exhibits strong robustness and generalization capability.

\begin{figure}[t!]
    \centering
    \subfigure[Effects on ACC]{
    \includegraphics[width=0.46\columnwidth]{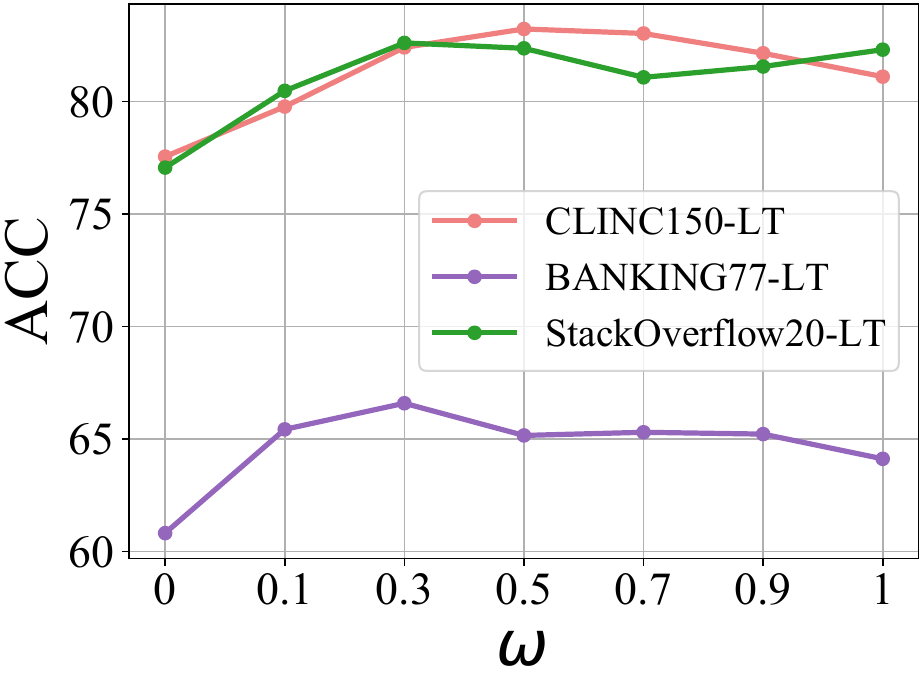}
    \label{tsne_3}
    }
    \subfigure[Effects on NMI]{
    \includegraphics[width=0.46\columnwidth]{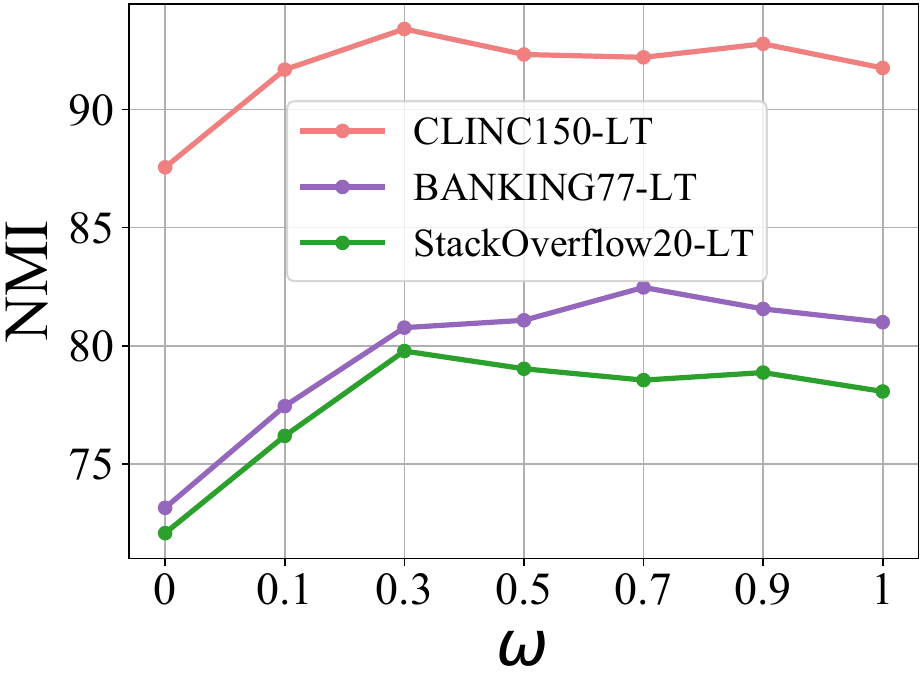}
    \label{tsne_4}
    }
    \caption{Effects of $\omega$ on~\dataset{}.} 
    \label{fig:t-SNE-1}
\end{figure}

\subsection{Effect of Exploration and Utilization} 
The weight of the multitask learning $\omega$ in Eq.~\ref{eq:final_loss} adjusts the contribution of two objectives.
Intuitively, the first term aims to explore cluster-friendly intent representations across all samples, while the second term focuses on mitigating the risk of catastrophic forgetting, ensuring the effective utilization of knowledge derived from clean samples.
We vary the value of $\omega$ and conduct experiments on~\dataset{} ($\gamma=10$) to explore the effect of $\omega$, which also reflects the inference of exploration and utilization.
In Fig.~\ref{fig:t-SNE-1}, only utilizing clean samples ($\omega=0.0$) or only exploring($\omega=1.0$) the intent representation will not achieve the best results. 
Interestingly, the effect of $\omega$ shows a similar trend (increase first and then decrease) on all metrics and datasets, which indicates that we can adjust the value of $\omega$ to give full play to the role of both so that the model can make better use of known knowledge to discover intents accurately.

\subsection{Comparison of Time Complexity}
\label{sex:appidx_computational_complexity}
The majority of existing methods~\cite{zhang-2022-new-intent-discovery,DPN,zhou2023latent} are mostly based on k-means for pseudo-labeling, while we propose a novel ROT approach for pseudo-labeling. 
We discuss the comparison and selection of time complexity between pseudo-labeling methods based on k-means and ROT.
Specifically, the k-means method is a clustering-based approach that iteratively computes distances between data points and assigns them to $k$ cluster centers. Its time complexity, typically around $O(nkt)$, depends on the dataset size ($n$), the number of cluster centers ($k$), and the convergence speed ($t$).
While the k-means method has lower time complexity, it is sensitive to the selection of initial cluster centers and convergence, leading to potentially unstable outcomes. On the other hand, ROT involves iteratively optimizing the distance or similarity between two data distributions to find the best mapping. Although the time complexity of ROT methods, such as those based on the Sinkhorn algorithm, is typically polynomial (e.g., $O(n^2m)$ where $n$ is the number of source domain data points and $m$ is the number of target domain data points), they generally provide high-quality pseudo-labels.

\section{Related Work}

\subsection{New Intent Discovery} 
New Intent Discovery (NID) is similar to generalized category discovery (GCD)~\cite{GCD}, which originates from computer vision and aims to discover novel intents by utilizing the prior knowledge of known intents.
\citet{lin2020discovering} conducts pair-wise similarity prediction to discover novel intents, and~\citet{zhang2021discovering} uses aligned pseudo-labels to help the model learn discriminative intent representations. 
Recent works further advance NID by incorporating contrastive learning~\cite{shen2021semi,NAACL2022,zhang-2022-new-intent-discovery,usnid}, knowledge transfer~\cite{DPN}, probabilistic frameworks~\cite{zhou2023latent}, pseudo-label learning~\cite{RoNID_zhang} or prototype attracting and dispersing~\cite{RAP_zhang} to capture cluster-friendly intent representation.
However, those methods operate under the unrealistic assumption that the distribution of both known and new intent classes is uniform, overlooking the long-tailed distributions frequently encountered in real-world scenarios. 
In this work, we explore the imbalanced NID scenario.

\subsection{Optimal Transport} 
Optimal Transport (OT) aims to find the most efficient transportation plan while adhering to marginal distribution constraints. 
It has been used in a broad spectrum of various tasks, including generative model~\cite{OT_2017_Ishaan}, semi-supervised learning~\cite{OT_2020_Fariborz,OT_2021_Kai}, clustering~\cite{OT_2020_Mathilde_Caron,OT_long_tail} and new intent discovery~\cite{RoNID_zhang}.
However, these methods typically impose an equality constraint when solving the OT problem. 
In contrast, we explore generating pseudo-labels by solving a relaxed OT problem. 
This approach encourages a uniform class distribution and addresses class degeneration in long-tailed 

\subsection{Contrastive Learning}
Contrastive Learning (CL) has been widely adopted to generate discriminative sentence representations for various scenarios~\cite{selfSCL,SCL2020,PCL_ICLR}, such as out-of-domain detection~\cite{DASFAA_zhang,ICASSP_zhang}, machine translation~\cite{alm,wmt2021,xmt,um4,m3p}, and named entity recognition~\cite{crop,mclner}. 
In essence, the primary intuition behind CL is to pull together positive pairs in the feature space while pushing away negative pairs. Motivated by its superior performance, contrastive learning has also been leveraged for intent recognition where it is used for NID. 
In this work, we design both class-wise and instance-wise contrastive learning objectives to learn cluster-friendly intent representations.

\section{Conclusion}
\label{sec:Conclusion}
In this work, we first propose the~\taskname{} task to identify known and infer novel intents within these long-tailed distributions.
Then, we develop an effective~\modelname{} baseline method for the~\taskname{} task, where pseudo-label generation and representation learning mutually iterate to achieve cluster-friendly representations.
Comprehensive experimental results on our~\dataset{} benchmark datasets demonstrate the effectiveness of our ~\modelname{} method for~\taskname{}.
We hope our work will draw more attention from the community toward a broader view of tackling the~\taskname{} problem.

\section*{Limitations}
\label{sec:limitations}
To better enlighten the follow-up research, we conclude the limitations of our method as follows:
(1) Enhancing interpretability. Our~\modelname{} automatically assigns labels to unlabeled utterances in real-world long-tail data distributions, yet it does not generate interpretable intent names for each cluster.
(2) Integration with LLMs. Large-scale language models (LLMs) have shown an impressive ability in a variety of NLP tasks, we plan to explore the integration of~\modelname{} with LLMs to boost performance in practical scenarios.
(3) Reducing time complexity. The time complexity of relaxed optimal transport (ROT) is $O\left(n^{2}\right)$, we plan to further develop a fast matrix scaling algorithm to reduce the complexity.

\section*{Acknowledgements}
\label{sec:Acknowledgements}
This work was supported in part by the National Natural Science Foundation of China (Grant Nos. U1636211, U2333205, 61672081, 62302025, 62276017), a fund project: State Grid Co., Ltd. Technology R\&D Project (ProjectName: Research on Key Technologies of Data Scenario-based Security Governance and Emergency Blocking in Power Monitoring System, Proiect No.: 5108-202303439A-3-2-ZN), the 2022 CCF-NSFOCUS Kun-Peng Scientific Research Fund and the Opening Project of Shanghai Trusted Industrial Control Platform and the State Key Laboratory of Complex \& Critical Software Environment (Grant No. SKLSDE-2021ZX-18).


\bibliography{anthology,custom}

\begin{thebibliography}{45}
\expandafter\ifx\csname natexlab\endcsname\relax\def\natexlab#1{#1}\fi

\bibitem[{An et~al.(2023)An, Tian, Zheng, Ding, Wang, and Chen}]{DPN}
Wenbin An, Feng Tian, Qinghua Zheng, Wei Ding, QianYing Wang, and Ping Chen. 2023.
\newblock Generalized category discovery with decoupled prototypical network.
\newblock In \emph{Proc. of AAAI}.

\bibitem[{Asano et~al.(2020)Asano, Rupprecht, and Vedaldi}]{naive_OT}
Yuki~Markus Asano, Christian Rupprecht, and Andrea Vedaldi. 2020.
\newblock Self-labelling via simultaneous clustering and representation learning.
\newblock In \emph{Proc. of ICLR}.

\bibitem[{Caron et~al.(2018)Caron, Bojanowski, Joulin, and Douze}]{caron2018deep}
Mathilde Caron, Piotr Bojanowski, Armand Joulin, and Matthijs Douze. 2018.
\newblock Deep clustering for unsupervised learning of visual features.
\newblock In \emph{Proc. of ECCV}.

\bibitem[{Caron et~al.(2020{\natexlab{a}})Caron, Misra, Mairal, Goyal, Bojanowski, and Joulin}]{OT_2020_Mathilde_Caron}
Mathilde Caron, Ishan Misra, Julien Mairal, Priya Goyal, Piotr Bojanowski, and Armand Joulin. 2020{\natexlab{a}}.
\newblock Unsupervised learning of visual features by contrasting cluster assignments.
\newblock In \emph{Proc. of NeurIPS}.

\bibitem[{Caron et~al.(2020{\natexlab{b}})Caron, Misra, Mairal, Goyal, Bojanowski, and Joulin}]{caron2020swav}
Mathilde Caron, Ishan Misra, Julien Mairal, Priya Goyal, Piotr Bojanowski, and Armand Joulin. 2020{\natexlab{b}}.
\newblock Unsupervised learning of visual features by contrasting cluster assignments.
\newblock In \emph{Proc. of NeurIPS}.

\bibitem[{Casanueva et~al.(2020)Casanueva, Tem{\v{c}}inas, Gerz, Henderson, and Vuli{\'c}}]{casanueva2020efficient}
I{\~n}igo Casanueva, Tadas Tem{\v{c}}inas, Daniela Gerz, Matthew Henderson, and Ivan Vuli{\'c}. 2020.
\newblock Efficient intent detection with dual sentence encoders.
\newblock In \emph{Proceedings of the 2nd Workshop on Natural Language Processing for Conversational AI}.

\bibitem[{Chen et~al.(2020)Chen, Kornblith, Norouzi, and Hinton}]{selfSCL}
Ting Chen, Simon Kornblith, Mohammad Norouzi, and Geoffrey~E. Hinton. 2020.
\newblock A simple framework for contrastive learning of visual representations.
\newblock In \emph{Proc. of ICML}.

\bibitem[{Chrabrowa et~al.(2023)Chrabrowa, Hadeliya, Kajtoch, Mroczkowski, and Rybak}]{nid_industry_setting}
Aleksandra Chrabrowa, Tsimur Hadeliya, Dariusz Kajtoch, Robert Mroczkowski, and Piotr Rybak. 2023.
\newblock Going beyond research datasets: Novel intent discovery in the industry setting.
\newblock In \emph{Proc. of ACL Findings}.

\bibitem[{Cui et~al.(2019)Cui, Jia, Lin, Song, and Belongie}]{imbalance_gama}
Yin Cui, Menglin Jia, Tsung{-}Yi Lin, Yang Song, and Serge~J. Belongie. 2019.
\newblock Class-balanced loss based on effective number of samples.
\newblock In \emph{Proc. of CVPR}.

\bibitem[{Cuturi(2013)}]{cuturi2013sinkhorn}
Marco Cuturi. 2013.
\newblock Sinkhorn distances: Lightspeed computation of optimal transport.
\newblock \emph{Proc. of NeurIPS}.

\bibitem[{Devlin et~al.(2019)Devlin, Chang, Lee, and Toutanova}]{Devlin2019BERTPO}
J.~Devlin, Ming-Wei Chang, Kenton Lee, and Kristina Toutanova. 2019.
\newblock Bert: Pre-training of deep bidirectional transformers for language understanding.
\newblock In \emph{Proc. of AACL}.

\bibitem[{Gulrajani et~al.(2017)Gulrajani, Ahmed, Arjovsky, Dumoulin, and Courville}]{OT_2017_Ishaan}
Ishaan Gulrajani, Faruk Ahmed, Mart{\'{\i}}n Arjovsky, Vincent Dumoulin, and Aaron~C. Courville. 2017.
\newblock Improved training of wasserstein gans.
\newblock In \emph{Proc. of NeurIPS}.

\bibitem[{Khosla et~al.(2020)Khosla, Teterwak, Wang, Sarna, Tian, Isola, Maschinot, Liu, and Krishnan}]{SCL2020}
Prannay Khosla, Piotr Teterwak, Chen Wang, Aaron Sarna, Yonglong Tian, Phillip Isola, Aaron Maschinot, Ce~Liu, and Dilip Krishnan. 2020.
\newblock Supervised contrastive learning.
\newblock In \emph{Proc. of NeurIPS}.

\bibitem[{Kumar et~al.(2022)Kumar, Patidar, Varshney, Vig, and Shroff}]{NAACL2022}
Rajat Kumar, Mayur Patidar, Vaibhav Varshney, Lovekesh Vig, and Gautam Shroff. 2022.
\newblock Intent detection and discovery from user logs via deep semi-supervised contrastive clustering.
\newblock In \emph{Proc. of NAACL}.

\bibitem[{Larson et~al.(2019)Larson, Mahendran, Peper, Clarke, Lee, Hill, Kummerfeld, Leach, Laurenzano, Tang, and Mars}]{larson2019clinc}
Stefan Larson, Anish Mahendran, Joseph~J. Peper, Christopher Clarke, Andrew Lee, Parker Hill, Jonathan~K. Kummerfeld, Kevin Leach, Michael~A. Laurenzano, Lingjia Tang, and Jason Mars. 2019.
\newblock An evaluation dataset for intent classification and out-of-scope prediction.
\newblock In \emph{Proc. of EMNLP}.

\bibitem[{Li et~al.(2021)Li, Zhou, Xiong, and Hoi}]{PCL_ICLR}
Junnan Li, Pan Zhou, Caiming Xiong, and Steven C.~H. Hoi. 2021.
\newblock Prototypical contrastive learning of unsupervised representations.
\newblock In \emph{Proc. of ICLR}.

\bibitem[{Li et~al.()Li, Dai, Simsek, Meinel, and Yang}]{imbagcd}
Ziyun Li, Ben Dai, Furkan Simsek, Christoph Meinel, and Haojin Yang.
\newblock Imbagcd: Imbalanced generalized category discovery.

\bibitem[{Lin et~al.(2020)Lin, Xu, and Zhang}]{lin2020discovering}
Ting-En Lin, Hua Xu, and Hanlei Zhang. 2020.
\newblock Discovering new intents via constrained deep adaptive clustering with cluster refinement.
\newblock In \emph{Proc. of AAAI}.

\bibitem[{Mo et~al.(2024)Mo, Yang, Liu, Wang, Chen, Wang, and Li}]{mclner}
Ying Mo, Jian Yang, Jiahao Liu, Qifan Wang, Ruoyu Chen, Jingang Wang, and Zhoujun Li. 2024.
\newblock {MCL-NER:} cross-lingual named entity recognition via multi-view contrastive learning.
\newblock In \emph{Proc. of AAAI}.

\bibitem[{Mou et~al.(2022)Mou, He, Wu, Wang, Wang, Wu, Huang, Feng, and Xu}]{gid}
Yutao Mou, Keqing He, Yanan Wu, Pei Wang, Jingang Wang, Wei Wu, Yi~Huang, Junlan Feng, and Weiran Xu. 2022.
\newblock Generalized intent discovery: Learning from open world dialogue system.
\newblock In \emph{Proc. of COLING}.

\bibitem[{Raedt et~al.(2023)Raedt, Godin, Demeester, and Develder}]{idas}
Maarten~De Raedt, Fr{\'{e}}deric Godin, Thomas Demeester, and Chris Develder. 2023.
\newblock {IDAS:} intent discovery with abstractive summarization.
\newblock \emph{CoRR}.

\bibitem[{Shen et~al.(2021)Shen, Sun, Zhang, and Najmabadi}]{shen2021semi}
Xiang Shen, Yinge Sun, Yao Zhang, and Mani Najmabadi. 2021.
\newblock Semi-supervised intent discovery with contrastive learning.
\newblock In \emph{Proceedings of the 3rd Workshop on Natural Language Processing for Conversational AI}.

\bibitem[{Shi et~al.(2023)Shi, An, Tian, Zheng, Wang, and Chen}]{DWGF}
Wenkai Shi, Wenbin An, Feng Tian, Qinghua Zheng, QianYing Wang, and Ping Chen. 2023.
\newblock A diffusion weighted graph framework for new intent discovery.
\newblock \emph{arXiv preprint arXiv:2310.15836}.

\bibitem[{Siddique et~al.(2021)Siddique, Jamour, Xu, and Hristidis}]{zero_shot_gid}
A.~B. Siddique, Fuad~T. Jamour, Luxun Xu, and Vagelis Hristidis. 2021.
\newblock Generalized zero-shot intent detection via commonsense knowledge.
\newblock In \emph{Proc. of SIGIR}.

\bibitem[{Taherkhani et~al.(2020)Taherkhani, Dabouei, Soleymani, Dawson, and Nasrabadi}]{OT_2020_Fariborz}
Fariborz Taherkhani, Ali Dabouei, Sobhan Soleymani, Jeremy~M. Dawson, and Nasser~M. Nasrabadi. 2020.
\newblock Transporting labels via hierarchical optimal transport for semi-supervised learning.
\newblock In \emph{Proc. of ECCV}.

\bibitem[{Tai et~al.(2021)Tai, Bailis, and Valiant}]{OT_2021_Kai}
Kai~Sheng Tai, Peter Bailis, and Gregory Valiant. 2021.
\newblock Sinkhorn label allocation: Semi-supervised classification via annealed self-training.
\newblock In \emph{Proc. of ICML}.

\bibitem[{Vaze et~al.(2022)Vaze, Han, Vedaldi, and Zisserman}]{GCD}
Sagar Vaze, Kai Han, Andrea Vedaldi, and Andrew Zisserman. 2022.
\newblock Generalized category discovery.
\newblock In \emph{Proc. of CVPR}.

\bibitem[{Xu et~al.(2015)Xu, Wang, Tian, Xu, Zhao, Wang, and Hao}]{xu2015short}
Jiaming Xu, Peng Wang, Guanhua Tian, Bo~Xu, Jun Zhao, Fangyuan Wang, and Hongwei Hao. 2015.
\newblock Short text clustering via convolutional neural networks.
\newblock In \emph{Proceedings of the 1st Workshop on Vector Space Modeling for Natural Language Processing}.

\bibitem[{Yang et~al.(2024)Yang, Guo, Yin, Bai, Wang, Liu, Liang, Cahi, Yang, and Li}]{m3p}
Jian Yang, Hongcheng Guo, Yuwei Yin, Jiaqi Bai, Bing Wang, Jiaheng Liu, Xinnian Liang, Linzheng Cahi, Liqun Yang, and Zhoujun Li. 2024.
\newblock m3p: Towards multimodal multilingual translation with multimodal prompt.
\newblock \emph{arXiv preprint arXiv:2403.17556}.

\bibitem[{Yang et~al.(2022{\natexlab{a}})Yang, Huang, Ma, Yin, Dong, Zhang, Guo, Li, and Wei}]{crop}
Jian Yang, Shaohan Huang, Shuming Ma, Yuwei Yin, Li~Dong, Dongdong Zhang, Hongcheng Guo, Zhoujun Li, and Furu Wei. 2022{\natexlab{a}}.
\newblock {CROP:} zero-shot cross-lingual named entity recognition with multilingual labeled sequence translation.
\newblock In \emph{Proc. of EMNLP Findings}.

\bibitem[{Yang et~al.(2023)Yang, Ma, Dong, Huang, Huang, Yin, Zhang, Yang, Wei, and Li}]{GanLM}
Jian Yang, Shuming Ma, Li~Dong, Shaohan Huang, Haoyang Huang, Yuwei Yin, Dongdong Zhang, Liqun Yang, Furu Wei, and Zhoujun Li. 2023.
\newblock Ganlm: Encoder-decoder pre-training with an auxiliary discriminator.
\newblock In \emph{Proc. of ACL}.

\bibitem[{Yang et~al.(2021{\natexlab{a}})Yang, Ma, Huang, Zhang, Dong, Huang, Muzio, Singhal, Hassan, Song, and Wei}]{wmt2021}
Jian Yang, Shuming Ma, Haoyang Huang, Dongdong Zhang, Li~Dong, Shaohan Huang, Alexandre Muzio, Saksham Singhal, Hany Hassan, Xia Song, and Furu Wei. 2021{\natexlab{a}}.
\newblock Multilingual machine translation systems from microsoft for {WMT21} shared task.
\newblock In \emph{Proceedings of the Sixth Conference on Machine Translation, WMT@EMNLP 2021, Online Event, November 10-11, 2021}.

\bibitem[{Yang et~al.(2020)Yang, Ma, Zhang, Wu, Li, and Zhou}]{alm}
Jian Yang, Shuming Ma, Dongdong Zhang, Shuangzhi Wu, Zhoujun Li, and Ming Zhou. 2020.
\newblock Alternating language modeling for cross-lingual pre-training.
\newblock In \emph{Proc. of AAAI}.

\bibitem[{Yang et~al.(2021{\natexlab{b}})Yang, Yin, Ma, Huang, Zhang, Li, and Wei}]{xmt}
Jian Yang, Yuwei Yin, Shuming Ma, Haoyang Huang, Dongdong Zhang, Zhoujun Li, and Furu Wei. 2021{\natexlab{b}}.
\newblock Multilingual agreement for multilingual neural machine translation.
\newblock In \emph{Proc. of ACL}.

\bibitem[{Yang et~al.(2022{\natexlab{b}})Yang, Yin, Ma, Zhang, Wu, Guo, Li, and Wei}]{um4}
Jian Yang, Yuwei Yin, Shuming Ma, Dongdong Zhang, Shuangzhi Wu, Hongcheng Guo, Zhoujun Li, and Furu Wei. 2022{\natexlab{b}}.
\newblock {UM4:} unified multilingual multiple teacher-student model for zero-resource neural machine translation.
\newblock In \emph{Proc. of IJCAI}.

\bibitem[{Zhang et~al.(2023{\natexlab{a}})Zhang, Xu, and He}]{OT_long_tail}
Chuyu Zhang, Ruijie Xu, and Xuming He. 2023{\natexlab{a}}.
\newblock Novel class discovery for long-tailed recognition.
\newblock \emph{CoRR}.

\bibitem[{Zhang et~al.(2021{\natexlab{a}})Zhang, Xu, Lin, and Lyu}]{zhang2021discovering}
Hanlei Zhang, Hua Xu, Ting-En Lin, and Rui Lyu. 2021{\natexlab{a}}.
\newblock Discovering new intents with deep aligned clustering.
\newblock In \emph{Proc. of AAAI}.

\bibitem[{Zhang et~al.(2021{\natexlab{b}})Zhang, Xu, Lin, and Lyu}]{zhang2021aligned}
Hanlei Zhang, Hua Xu, Ting-En Lin, and Rui Lyu. 2021{\natexlab{b}}.
\newblock Discovering new intents with deep aligned clustering.
\newblock \emph{Proc. of AAAI}.

\bibitem[{Zhang et~al.(2023{\natexlab{b}})Zhang, Xu, Wang, Long, and Gao}]{usnid}
Hanlei Zhang, Hua Xu, Xin Wang, Fei Long, and Kai Gao. 2023{\natexlab{b}}.
\newblock {USNID:} {A} framework for unsupervised and semi-supervised new intent discovery.
\newblock \emph{CoRR}.

\bibitem[{Zhang et~al.(2023{\natexlab{c}})Zhang, Bai, Li, Yan, and Li}]{DASFAA_zhang}
Shun Zhang, Jiaqi Bai, Tongliang Li, Zhao Yan, and Zhoujun Li. 2023{\natexlab{c}}.
\newblock Modeling intra-class and inter-class constraints for out-of-domain detection.
\newblock In \emph{Proc. of DASFAA}.

\bibitem[{Zhang et~al.(2023{\natexlab{d}})Zhang, Li, Bai, and Li}]{ICASSP_zhang}
Shun Zhang, Tongliang Li, Jiaqi Bai, and Zhoujun Li. 2023{\natexlab{d}}.
\newblock Label-guided contrastive learning for out-of-domain detection.
\newblock In \emph{Proc. of ICASSP}.

\bibitem[{Zhang et~al.(2024{\natexlab{a}})Zhang, Yan, Yang, Ren, Bai, Li, and Li}]{RoNID_zhang}
Shun Zhang, Chaoran Yan, Jian Yang, Changyu Ren, Jiaqi Bai, Tongliang Li, and Zhoujun Li. 2024{\natexlab{a}}.
\newblock Ronid: New intent discovery with generated-reliable labels and cluster-friendly representations.
\newblock \emph{CoRR}.

\bibitem[{Zhang et~al.(2024{\natexlab{b}})Zhang, Yang, Bai, Yan, Li, Yan, and Li}]{RAP_zhang}
Shun Zhang, Jian Yang, Jiaqi Bai, Chaoran Yan, Tongliang Li, Zhao Yan, and Zhoujun Li. 2024{\natexlab{b}}.
\newblock New intent discovery with attracting and dispersing prototype.
\newblock \emph{arXiv preprint arXiv:2403.16913}.

\bibitem[{Zhang et~al.(2022)Zhang, Zhang, Zhan, Wu, and Lam}]{zhang-2022-new-intent-discovery}
Yuwei Zhang, Haode Zhang, Li-Ming Zhan, Xiao-Ming Wu, and Albert Lam. 2022.
\newblock New intent discovery with pre-training and contrastive learning.
\newblock In \emph{Proc. of ACL}.

\bibitem[{Zhou et~al.(2023)Zhou, Quan, and Qiu}]{zhou2023latent}
Yunhua Zhou, Guofeng Quan, and Xipeng Qiu. 2023.
\newblock A probabilistic framework for discovering new intents.
\newblock In \emph{Proc. of ACL}.

\end{thebibliography}

\appendix

\begin{table*}[!ht]
    \centering
    \resizebox{0.8\textwidth}{!}{
    \begin{tabular}{l|c|c|c|c|c|c}
    \toprule
    Dataset & Classes & \#Training & \#Validation & \#Testing & Vocabulary & Length (Max / Avg) \\
    \midrule
     CLINC & 150 & 18000 & 2250 & 2250 & 7283 & 28 / 8.32  \\
     BANKING & 77 & 9003 & 1000 & 3080 & 5028 & 79 / 11.91 \\
     StackOverflow & 20 & 12000 & 2000 & 1000 & 17182 & 41 / 9.18 \\
    \bottomrule
    \end{tabular}
    }
    \caption{Statistics of original datasets. \# denotes the total number of utterances.}
    \label{tab:origin_three_datasets}
\end{table*}
\begin{figure*}[t]
    \centering
    \subfigure[BANKING77-LT]{
    \includegraphics[width=0.92\columnwidth]{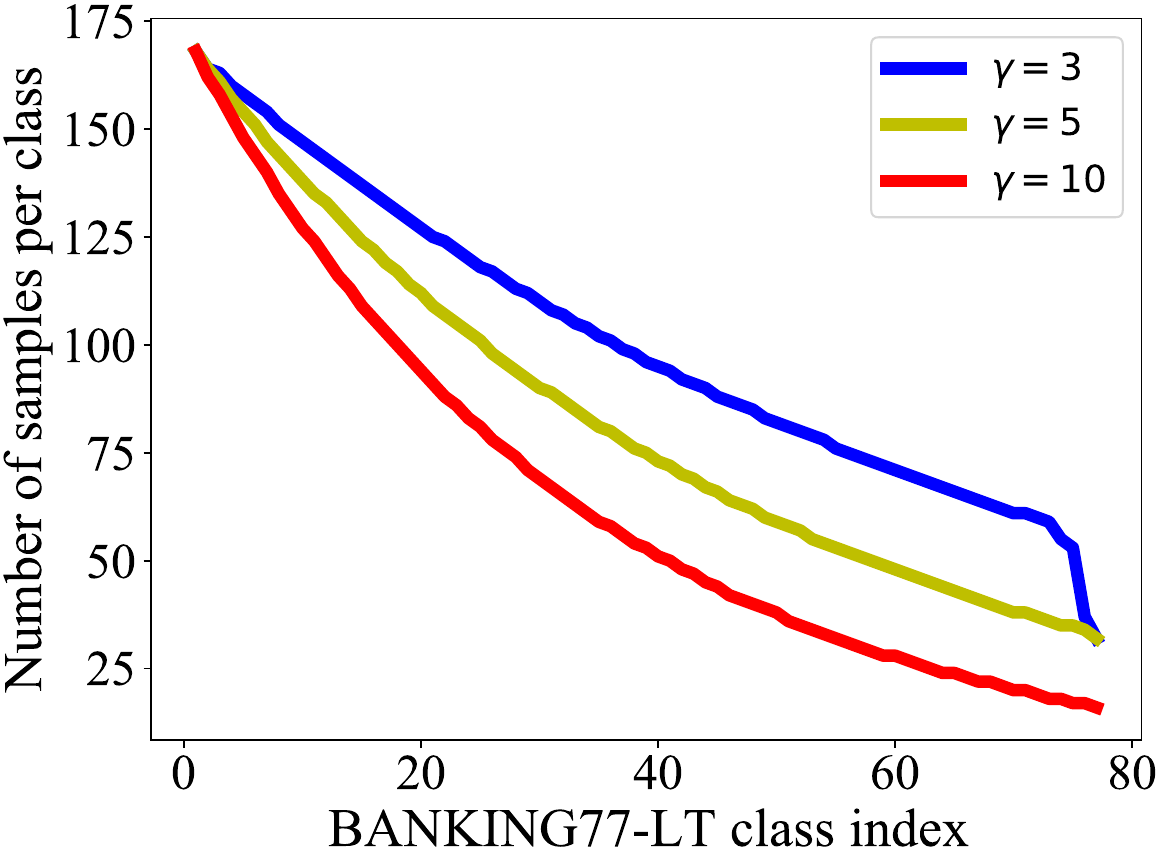}
    \label{imbalancd_banking_77_LT}
    }
    \subfigure[StackOverflow20-LT]{
    \includegraphics[width=0.9\columnwidth]{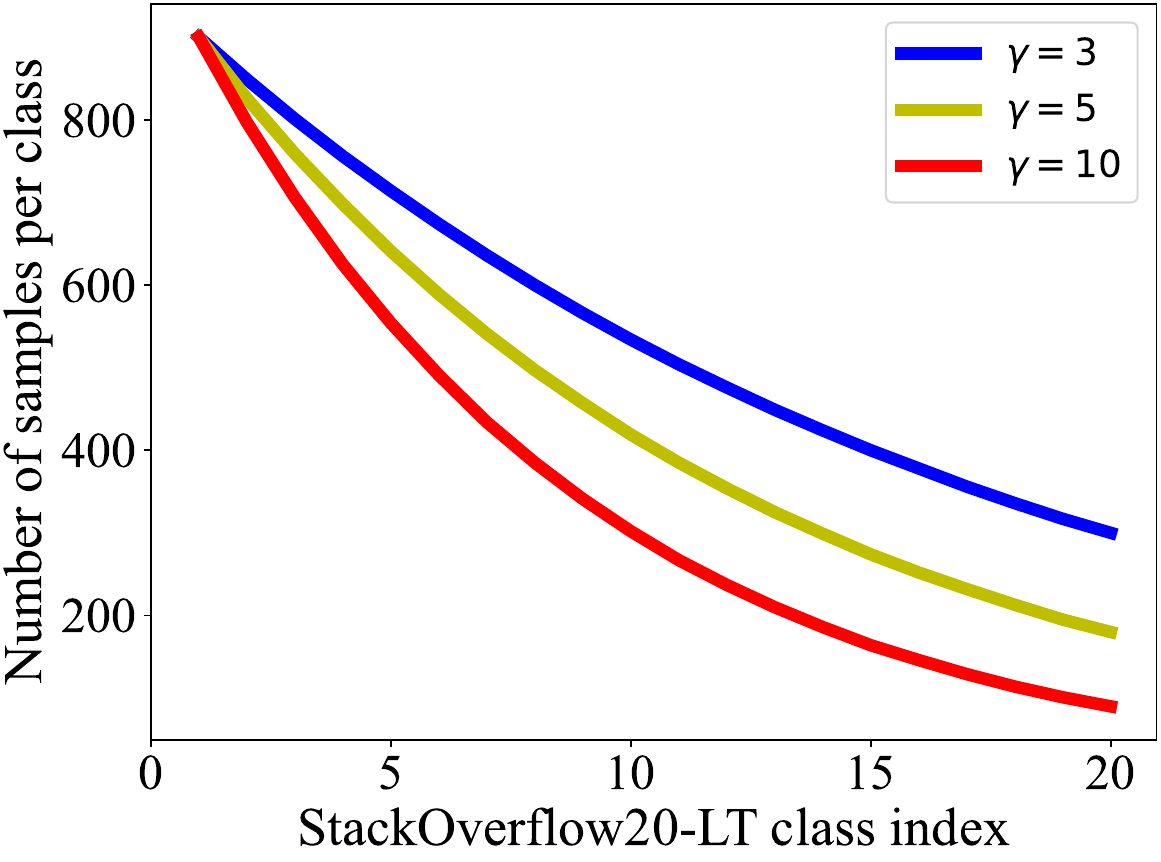}
    \label{imbalance_stack20_LT}
    }
    \vspace{-5pt}
    \caption{Number of training samples per class in artificially created long-tailed BANKING77-LT and StackOverflow20-LT datasets with different imbalance factors.} 
    \vspace{-5pt}
    \label{fig:appdix_imbalance}
\end{figure*}
%
\begin{table}[t]
\centering
    \resizebox{1.0\columnwidth}{!}{
    \begin{tabular}{l|c|c|c|c|c}
    \toprule
    \dataset{} ($\gamma=3$) & $|\mathcal{Y}^{k}|$  & $|\mathcal{Y}^{n}|$ & $|\mathcal{D}_{l}|$ & $|\mathcal{D}_{u}|$ & $|\mathcal{D}_{t}|$ \\
    \midrule
     CLINC150-LT & 113 & 37 & 868 & 9995 & 2250  \\
     BANKING77-LT & 58 & 19 & 607 & 7163 & 3080  \\
     StackOverflow20-LT & 15 & 5 & 830 & 10140 & 1000  \\
     \midrule
     \midrule
     \dataset{} ($\gamma=5$) & $|\mathcal{Y}^{k}|$  & $|\mathcal{Y}^{n}|$ & $|\mathcal{D}_{l}|$ & $|\mathcal{D}_{u}|$ & $|\mathcal{D}_{t}|$ \\
     \midrule
     CLINC150-LT & 113 & 37 & 719 & 8164 & 2250  \\
     BANKING77-LT & 58 & 19 & 487 & 5924 & 3080  \\
     StackOverflow20-LT & 15 & 5 & 686 & 8350 & 1000  \\
    \bottomrule
    \end{tabular}}
    \caption{Statistics of the~\dataset{} datasets when $\gamma=3$ and $\gamma=5$. 
    $|\mathcal{Y}^{k}|$, $|\mathcal{Y}^{n}|$,  $|\mathcal{D}_{l}|$,  $|\mathcal{D}_{u}|$ and $|\mathcal{D}_{t}|$ represent the number of known categories, novel categories, labeled data, unlabeled data, and testing data.}
    \label{tab:imbalance_3_5}
\end{table}

\section{ROT}
\label{sec:ROT_optimal}
In this section, we provide a comprehensive optimization process for the ROT problem~\eqref{eq:OT_KL}, the ROT objective is:

\begin{BigEquation}
\begin{equation}
    \begin{aligned}
        &\min_{\mathbf{Q},\bm{\beta}}{\langle \mathbf{Q}, {-\log{\mathbf{P}}} \rangle} + \lambda_1 H(\mathbf{Q}) + \lambda_2 D_{\text{KL}}(\frac{1} {K}  \bm{1} , \bm{\beta})\\
        &s.t.\,\,\mathbf{Q}\bm{1}=\bm{\alpha},\mathbf{Q}^T\bm{1}=\bm{\beta}, \mathbf{Q}\geq0, \bm{\beta}^T\bm{1}=1
    \end{aligned}
\end{equation}
\end{BigEquation}where $\lambda_1$ and $\lambda_2$ are hyper-parameters, and $D_{\text{KL}}(\bm{A}, \bm{B})$ denotes the Kullback-Leibler Divergence. We utilize the Lagrangian multiplier algorithm for optimization:
\begin{BigEquation}
\begin{equation}
\label{eq:lq}
    \begin{aligned}
        L(\mathbf{Q}, \bm{\beta}, \bm{f}, \bm{g}, h)&={\langle \mathbf{Q}, {-\log{\mathbf{P}}} \rangle} + \lambda_1 H(\mathbf{Q}) \\
        &+ \lambda_2 D_{\text{KL}}(\frac{1} {K} \bm{1}, \bm{\beta}) -\bm{f}^T(\mathbf{Q}\bm{1}-\bm{\alpha})\\
        &-\bm{g}^T(\mathbf{Q}^T\bm{1}-\bm{\beta}) -h(\bm{\beta}^T\bm{1}-1)
    \end{aligned}
\end{equation}
\end{BigEquation}where $\bm{f}$, $\bm{g}$, and $h$ are Lagrangian multipliers.
Differentiating Eq.~(\ref{eq:lq}) yields the following result:
\begin{BigEquation}
\begin{equation}
\label{eq:pipd}
    \begin{aligned}
        \frac{\partial L}{\partial Q_{ij}}=\lambda_1 log(Q_{ij})-log(P_{ij})-f_{i}-g_{j}
    \end{aligned}
\end{equation}
\begin{equation}
\label{eq:fpd}
    \begin{aligned}
        \frac{\partial L}{\partial f_{i}}= -(\sum_{j}^{K} Q_{ij}) + \alpha_{i}
    \end{aligned}
\end{equation}
\begin{equation}
\label{eq:gpd}
    \begin{aligned}
        \frac{\partial L}{\partial g_{j}}= -(\sum_{i}^{N} Q_{ij}) + \beta_{j}
    \end{aligned}
\end{equation}
\begin{equation}
\label{eq:bpd}
    \begin{aligned}
        \frac{\partial L}{\partial \beta_{j}}=-\frac{\lambda_2}{K \beta_{j}}+g_{j}-h
    \end{aligned}
\end{equation}
\begin{equation}
\label{eq:hpd}
    \begin{aligned}
        \frac{\partial L}{\partial h}=-(\sum_{j}^{K} \beta_{j}) + 1
    \end{aligned}
\end{equation}
\end{BigEquation}Initially, we fix $\bm{\beta}$ and $h$, and then update $\mathbf{Q}$, $\bm{f}$, and $\bm{g}$. By setting $\frac{\partial L}{\partial Q_{ij}}$, $\frac{\partial L}{\partial f_{i}}$, and $\frac{\partial L}{\partial g_{j}}$ to zero, we obtain the following results:
\begin{BigEquation}
\begin{equation}
\label{eq:piijval}
    \begin{aligned}
        Q_{ij} &= \exp(\frac{f_{i}+\log(P_{ij})+g_{j}}{\lambda_1})\\
        &= \exp(\frac{f_{i}}{\lambda_1}) \cdot \exp(\frac{\log(P_{ij})}{\lambda_1}) \cdot \exp(\frac{g_{j}}{\lambda_1})
    \end{aligned}
\end{equation}
\begin{equation}
\label{eq:aconst}
    \begin{aligned}
        \sum_{j}^{K} Q_{ij} = \alpha_{i},
        \sum_{i}^{N} Q_{ij} = \beta_{j}
    \end{aligned}
\end{equation}
Based on Eq.~(\ref{eq:piijval}), we derive the following:
\begin{equation}
\label{eq:pival}
    \begin{aligned}
        \mathbf{Q} = \text{diag}(\exp(\frac{\bm{f}}{\lambda_1}))\exp(\frac{{\log{\mathbf{P}}}}{\lambda_1})\text{diag}(\exp(\frac{\bm{g}}{\lambda_1}))
    \end{aligned}
\end{equation}
\end{BigEquation}Considering the constraints~(\ref{eq:aconst}) and the conditions $\bm{\beta}^T\bm{1}=\bm{\alpha}^T\bm{1}=1$, we solve Eq.~(\ref{eq:pival}) to determine the values of $\mathbf{Q}$, $\bm{f}$, and $\bm{g}$ using the Sinkhorn algorithm~\cite{{cuturi2013sinkhorn}}. 
Subsequently, with $\bm{f}$, $\bm{g}$, and $\mathbf{Q}$ fixed, we update $\bm{\beta}$ and $h$. 
Setting Eq.~(\ref{eq:bpd}) to zero yields the following solution:
\begin{BigEquation}
\begin{equation}
\label{eq:btoh}
    \begin{aligned}
        \beta_{j} = \frac{\lambda_2}{K(g_j-h)}
    \end{aligned}
\end{equation}
\end{BigEquation}Take Eq.~(\ref{eq:btoh}) into the Eq.~(\ref{eq:hpd}) and let Eq.~(\ref{eq:hpd}) equal to 0, we can obtain:
\begin{BigEquation}
\begin{equation}
\label{eq:hequation}
    \begin{aligned}
        (\sum_{j}^{K} \beta_{j}(h)) - 1 = 0
    \end{aligned}
\end{equation}
\end{BigEquation}We obtain $h$ from Eq.(\ref{eq:hequation}) using the bisection method and subsequently determine the corresponding $\bm{\beta}$.\
In the final step, we iteratively update $\bm{f}$, $\bm{g}$, $\mathbf{Q}$, and $\bm{\beta}$, $h$. The iterative optimization process for ROT is outlined in Algorithm\ref{alg:rot}.
\begin{algorithm}[t]
\caption{The optimization of ROT }\label{alg:rot}
\textbf{Input:} The cost matrix: ${-\log{\mathbf{P}}}$.\\
\textbf{Output:} \\
The transport matrix: $\mathbf{Q}$,\\
The class distribution: $\bm{\beta}$.\\
\textbf{Procedure}:
\begin{algorithmic}[1]
{
\STATE{Initialize $\bm{\beta}$ as uniform distribution;}
\FOR{$i=1$ to $T$}
\STATE{Fix $\bm{\beta}$ and $h$, calculate $\mathbf{Q}$, $\bm{f}$ and $\bm{g}$ with Sinkhorn algorithm.}
\STATE{Fix $\mathbf{Q}$, $\bm{f}$ and $\bm{g}$, update $\bm{\beta}$ and $h$ with Eq.~(\ref{eq:btoh}) and~(\ref{eq:hequation}).
}
\ENDFOR
\STATE{Return $\mathbf{Q}$ and $\bm{\beta}$.}
}
\end{algorithmic}
\end{algorithm}

\section{Statistics of Datasets}
\label{sec:appendix_Datasets}
We present detailed statistics of the CLINC~\cite{larson2019clinc}, BANKING~\cite{casanueva2020efficient} and StackOverflow~\cite{xu2015short} datasets in Table~\ref{tab:origin_three_datasets}. 
In addition, we display the number of samples per class for BANKING77-LT and StackOverflow20-LT under various imbalance factors, as shown in Fig.~\ref{fig:appdix_imbalance}.
We also provide dataset statistics for the~\dataset{} datasets with imbalance factors of 3 and 5, as shown in Table~\ref{tab:imbalance_3_5}.
\section{Comparison Methods}
\label{sec:appendix_Comparison}
In this work, we compare the proposed~\modelname{} method against several representative baselines including:

\noindent
\textbf{GCD}~\cite{GCD} introduces a combination of supervised and self-supervised contrastive learning to learn distinctive representations, which are then clustered using k-means.

\noindent
\textbf{DeepAligned}~\cite{zhang2021discovering} is an improved DeepClustering~\cite{caron2018deep} that uses an alignment strategy to alleviate the label inconsistency problem.

\noindent
\textbf{MTP-CLNN}~\cite{zhang-2022-new-intent-discovery} is a method that applies multi-task pre-training and nearest neighbors contrastive learning for NID.

\noindent
\textbf{DPN}~\cite{DPN} proposes a decoupled prototypical network that, by framing a bipartite matching problem for category prototypes, separates known and novel categories to meet their distinct training objectives and transfers category-specific knowledge for capturing high-level semantics.

\noindent
\textbf{LatentEM}~\cite{zhou2023latent} introduces a principled probabilistic framework optimized with the EM algorithm.
In the E-step, it assigns pseudo-labels, and in the M-step, it learns cluster-friendly representations and updates parameters through contrastive learning.

\noindent
\textbf{USNID}~\cite{usnid} is a two-stage framework for both unsupervised and semi-supervised NID with an efficient centroid-guided clustering mechanism.

\section{Implementation Details}
\label{sec:appendix_Implementation}
To ensure a fair comparison for~\modelname{} and all baselines, we consistently adopt the pre-trained 12-layer bert-uncased BERT model\footnote{\url{https://huggingface.co/bert-base-uncased}} \cite{Devlin2019BERTPO} as the backbone encoder in all experiments and only fine-tune the last transformer layer parameters to expedite the training process as suggested in \cite{zhang2021discovering}.
We adopt the AdamW optimizer with 0.01 weight decay and 1.0 gradient clipping for parameter update.
During pre-training, we set the learning rate to 5e-5 and adopt the early stopping strategy with a patience of 20 epochs.
For CLNN~\cite{zhang-2022-new-intent-discovery}, the external dataset is not used as in other baselines, the parameter of top-k nearest neighbors is set to $\{$100, 50, 500$\}$ for CLINC, BANKING, and StackOverflow, respectively, as utilized in~\citet{zhang-2022-new-intent-discovery}. 
For all experiments, we set the batch size as 512 and the temperature scale as $\tau$ = 0.1 in Eq.~\eqref{eq:class_wise_CL} and Eq.~\eqref{eq:instance-wise_CL}.
We set the parameter $\rho$ = 0.65 in Eq.~\eqref{eq:distribution-aware}, the confidence threshold $\tau_{g}$ = 0.9 in Eq.~\eqref{eq:quality-aware}.
We adopt the data augmentation of random token replacement as~\citet{zhang-2022-new-intent-discovery}.
All experiments are conducted on 4 Tesla V100 GPUs and averaged over 3 runs.
we split the datasets into train, valid, and test sets, and randomly select 25\% of categories as unknown and only 10\% of training data as labeled. 
The number of intent categories is set as ground truth.



\section{Estimate the Number of Intents ($K$)}
\label{sex:appidx_estimate_K}
In practical dialogue systems, new intents emerge constantly and we cannot know the exact number of the intent clusters. 
In this paper, following the work of~\citep{zhang2021aligned}, we take the full usage of the well-initialized intent features to automatically estimate the intent cluster number $K$. 
Specifically, we first assign a big  $K^{\prime}$  as the initial intent cluster number. 
Then we directly use the pre-trained model to extract the feature representations for the training data and perform the K-means algorithm to group these feature representations into different clusters. 
From these clusters, we can distinguish the dense and boundary-clear clusters as the real intent clusters, while the remaining low-size clusters are filtered out. 
The filtering function can be formulated as follows:
\begin{BigEquation}
\begin{equation}
K=\sum_{i=1}^{K^{\prime}} \delta\left(\left|T_{i}\right| \geq t\right)
\end{equation}
\end{BigEquation}where $\left|T_{i}\right|$ is the size the  $i_{th}$  grouped cluster, $t$ is the threshold of filtering.  
$\delta(\cdot)$ is the indicator function, whose output is $1$ if the condition is satisfied.

\begin{figure}[t!]
    \centering
    \subfigure[Balanced]{
    \includegraphics[width=0.46\columnwidth]{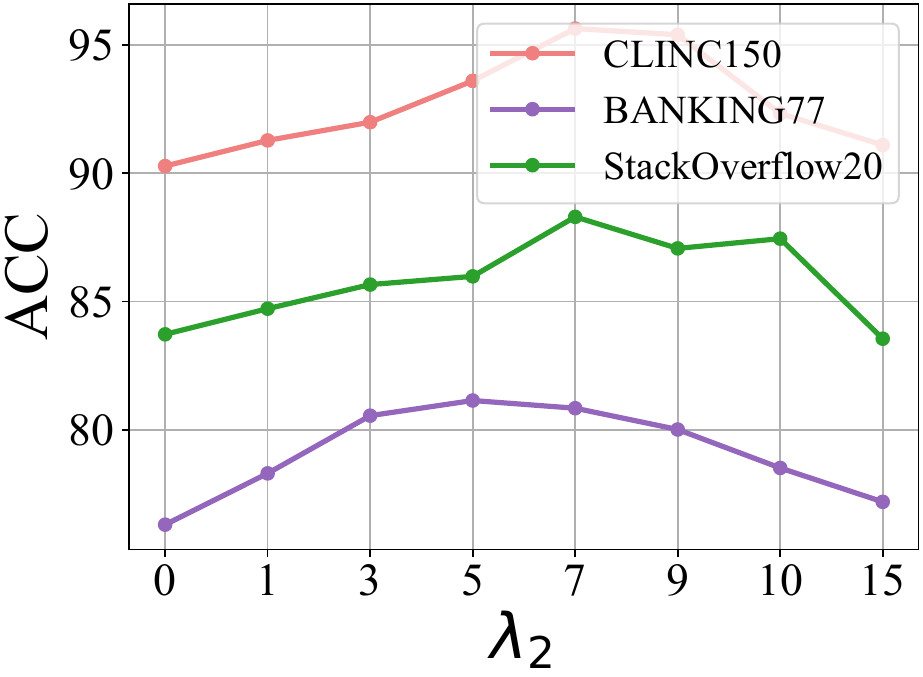}
    \label{tsne_3_balanced}
    }
    \subfigure[Imbalanced]{
    \includegraphics[width=0.46\columnwidth]{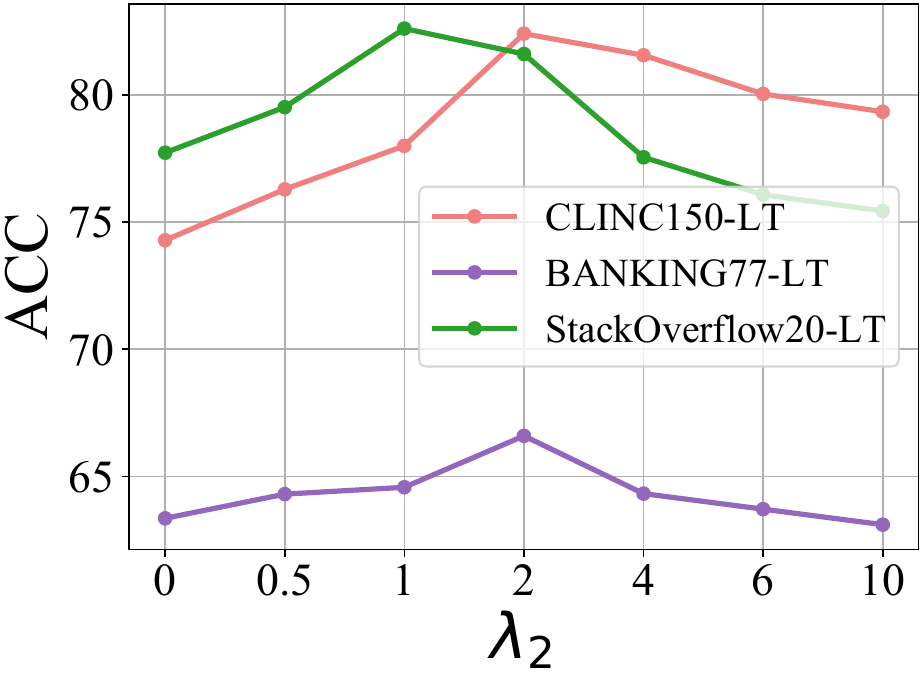}
    \label{tsne_3_imbalanced}
    }
    \caption{Effects of $\lambda_{2}$ on~\dataset{}.} 
    \label{fig:imbalanced}
    \vspace{-10pt}
\end{figure}

\section{Hyper-Parameter Analyses}
To investigate the sensitiveness of the hyper-parameters in Eq.~\ref{eq:OT_KL}, we first referred to the experience from previous studies~\cite{naive_OT,caron2020swav} and identified $\lambda_{1}=0.05$ on the all datasets.
Then we examine the impact of $\lambda_2$ on model performance by varying the value of $\lambda_2$ to observe the performance changes.
The results are reported in Fig.~\ref{fig:imbalanced}.
Specifically, Fig.~\ref{tsne_3_balanced} shows the impact of $\lambda_2$ variation on the performance of balanced datasets, while Fig.~\ref{tsne_3_imbalanced} demonstrates the effect of $\lambda_2$ on the performance of imbalanced datasets.
Empirically, we choose $\lambda_2=7$ on the balanced datasets, and $\lambda_2=2$ on the imbalanced~\dataset{} datasets.


\end{document}